# Cutting AI Datacenter Energy with Reinforcement Learning: Measured Power Control of LLM Training from One GPU to the Fleet

**Eliseo Curcio**

Reinforcement-learning post-training dominates modern language-model development, yet its power behavior on GPU hardware has not been characterized, and datacenters manage GPU power only with workload-blind mechanisms, static caps and reactive throttling, that slow the hardware indiscriminately. We instrument GRPO training with half-second power telemetry at 7B, 14B, and 72B scales on one to four A100s, over 380,000 samples, and train a PPO meta-controller that adapts the workload's own generation parameters to measured power. Evaluated against the full 500-step 7B trace, the controller cuts power-limit violations by 89.8% while increasing token output by 18.1% and energy efficiency by 26.2% in tokens per megawatt-hour. Deployed live at 72B, the same controller family yields replicated null results, diagnosed as the group-size actuator losing authority under model sharding. An actuator-authority sweep shows the same parameters applied as generation concurrency retain 17 to 22% power authority, isolating an occupancy-versus-volume principle; a controller rebuilt on that actuator controls a live 72B rollout-generation workload across three hardware replications: 35.7% more output than a static safe baseline at 2.27 ± 1.08% budget violations, 87.2% fewer violations than uncontrolled operation, and the highest mean throughput and lowest mean energy per token among the constrained controllers, with an adaptive threshold rule matching it in one of three operating conditions. Under realistic measurement windows the original 72B transients collapse from 23.6% at half-second resolution to 1.6% at thirty seconds and zero at five minutes; a composed sixteen-GPU fleet shows zero violations at thirty seconds and longer, with peak demand at 50 to 56% of nameplate. For this fleet mix, approximately twofold oversubscription of nameplate appears feasible for similar workloads, subject to operator validation. We quantify the economic and carbon consequences and specify a low-cost operator pilot to verify the estimate.

Advanced Department of Artificial Intelligence and Energy – New York

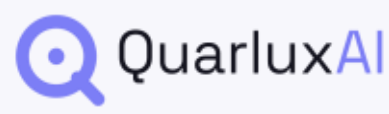

## 1. Introduction

The constraint that paces artificial-intelligence infrastructure has shifted from silicon supply to electricity. United States datacenter consumption reached an estimated 176 TWh in 2023, some 4.4% of national demand, and is projected to double or triple by 2028, driven principally by AI accelerators [1]. The International Energy Agency projects global datacenter consumption approaching 945 TWh by 2030 with artificial intelligence as the dominant growth driver [2], investment analyses forecast a 160% increase in datacenter power demand over this decade [3], and near-term electricity outlooks identify datacenters as the fastest-growing class of new demand in advanced economies [4]. The grid cannot absorb this growth on the industry's schedule. The United States interconnection queue held over 2,600 GW of proposed capacity in 2023, with typical waits of five years or more [5], federal regulatory reform explicitly targets the queue as a bottleneck [6], and utility planning bodies now treat datacenter load growth as the defining resource-adequacy question of the decade [7]. Independent analyses caution that the energy footprint of artificial intelligence is growing faster than any efficiency trend offsets it [8].

Production power management today is workload-blind: nameplate provisioning, firmware caps that clamp clocks at a static limit, thermal throttling that reacts to temperature. None knows what the computation is doing, so all buy safety by slowing everything, critical gradient synchronization and elastic generation alike. Reinforcement learning inverts this: instead of limiting hardware beneath the workload, a learned policy adapts the workload itself, spending the budget where it buys output and yielding it where it does not. Nothing is clamped; the job breathes under the budget. This distinction, hardware throttling against workload adaptation, is the difference the paper measures. For clusters the difference is decisive: a static cap must be set for the worst case; a learned policy tracks each moment.

Against this backdrop, the practice by which AI facilities purchase electrical capacity has escaped empirical scrutiny. Facilities are provisioned to nameplate: the sum of the thermal design power ratings of the installed accelerators, plus a safety margin. The practice was inherited from enterprise computing, where individual loads were small, mutually uncorrelated, and uncontrollable, so that the cost of a conservative sizing rule was modest. None of those three premises has been tested for modern AI training loads, and industry surveys already report rack-level utilization persistently below design assumptions [9]. Two considerations suggest the premises fail. The first is statistical: training jobs sharing a facility do not reach peak power simultaneously, so aggregate demand should fall below the sum of individual peaks, by the same diversity principle through which electric utilities have sized distribution networks for a century. The second is newer and specific to the current era of model development: reinforcement-learning post-training exposes runtime parameters on its dominant, power-hungry phase, which raises the possibility that training is not merely a statistical load but a controllable one, in utility terms a candidate dispatchable resource. Grid-integration research estimates that modest curtailability in large new loads would allow the existing United States grid to absorb roughly 100 GW of additional demand without new generation [10], and the Department of Energy identifies load flexibility as a primary lever for accommodating AI growth [11]. What has been missing from both literatures is measured evidence that AI training tolerates power control without sacrificing the output it exists to produce.

This paper supplies that evidence, from single accelerator to facility. We instrument reinforcement-learning training with fine-grained power telemetry at three model scales, build and evaluate a learned controller over the training loop's own parameters, locate the scale boundary at which such control stops working and explain why, re-examine what a power violation even means under the measurement windows real electrical infrastructure uses, and compose the measured traces into a facility-scale analysis of how much capacity a training fleet actually requires. The contributions are as follows. First, the paper provides what is to our knowledge the first power characterization of reinforcement-learning post-training, comprising approximately 200,000 GPU telemetry samples at half-second resolution in the original BF16 campaign across the 7B, 14B, and 72B scales, and over 380,000 samples in total once the generation-proxy campaign is included, with training phases labeled. Second, it demonstrates a reinforcement-learning power controller that, evaluated on the full measured 7B trace, reduces power-limit violations by 89.8% while increasing output by 18.1% and energy efficiency by 26.2%. Third, it establishes a measured scale boundary for in-loop control, replicated across three controllers on multi-GPU hardware, diagnoses its mechanism, and then crosses it: an actuator-authority sweep isolates generation concurrency as the software knob that retains 17 to 22% power authority under sharding, and a controller rebuilt on that occupancy actuator, trained with constrained reinforcement learning on measured transition traces, demonstrates live control of a 72B rollout-generation workload across three hardware replications with 35.7% more output than a static safe baseline. Fourth, it contributes a measurement finding: the power violations that remain at multi-GPU scale are sub-second transients largely suppressed under infrastructure-relevant averaging and eliminated by the five-minute window in the measured trace. Fifth, it derives from the measured traces a fleet-level provisioning analysis showing mixed-fleet peak demand of 50 to 56% of nameplate, and it quantifies the economic and carbon consequences of closing that gap.

## 2. Related work

Research on the power behavior of large language models has concentrated on inference. Patel et al. profiled power management opportunities in cloud LLM serving and demonstrated substantial oversubscription headroom in inference fleets [12], and the same group's Splitwise separated the compute-intensive prompt phase from the memory-bound token-generation phase onto different machines to exploit their distinct power profiles [13]. DynamoLLM extended this line to whole-cluster design, reconfiguring inference clusters for energy efficiency under latency constraints [14], while Samsi et al. benchmarked inference energy systematically across models and hardware generations [15]. These studies establish that serving workloads possess exploitable power structure. Training, however, differs from serving in loop structure, in the parameters available at runtime, and in the timescales of its power disturbances, and none of these systems addresses it.

A second line of work optimizes the energy of training itself. Zeus navigates the trade-off between GPU power limits and batch size to minimize the energy of deep-network training [16], and Perseus plans per-stage GPU frequencies through the computation graph of large-model training to remove what its authors call energy bloat [17]. Both systems target conventional supervised training, act through frequency or power-limit plans computed before execution, and treat the workload itself as fixed. Neither instruments reinforcement-learning post-training, and neither closes a feedback loop from live power telemetry back into the training algorithm's own parameters, which is the mechanism investigated here. A related accounting literature measures what models cost energetically, including the training footprint of a 176B-parameter model [18] and the deployment cost of models across tasks [19]; these studies motivate control but do not provide it. At the facility scale, recent industry work addresses the power oscillations that large synchronized training jobs impose on datacenter electrical systems, approaching the problem from the hardware and ramp-rate side [20]; that work is complementary to the software-side control developed in this paper.

On the methodological side, the controller developed here draws on established reinforcement-learning practice. Proximal Policy Optimization is the workhorse policy-gradient algorithm of language-model alignment [21], and our implementation uses standard open tooling [22, 23]. The constrained variant follows the Lagrangian approach surveyed in the safe reinforcement-learning literature [24], and the signal-protection variant applies constraint-by-construction through action-space bounds, a pattern formalized in contemporary safety benchmarks [25]. The configuration itself is to our knowledge novel: a reinforcement-learning agent governing the runtime parameters of another reinforcement-learning process, live, on hardware. The closest work in spirit is Meta's Autodata, an agentic system that optimizes training data under a compute budget while protecting learnability through a hard solver-gap criterion [26]. The present work can be read as the infrastructure analogue, optimizing power under a facility budget while protecting the training signal, and Section 4 shows that the two efforts converged independently on the same design lesson regarding how such protection must be implemented.

No prior work, to our knowledge, instruments the power behavior of reinforcement-learning post-training, closes a live control loop over its parameters, or connects measured training power behavior to facility provisioning practice. This paper addresses all three.

## 3. Background

Large language models are trained in stages. Pretraining on large corpora, at compute scales that have grown relentlessly with each model generation [27], produces general linguistic capability but not aligned or purposeful behavior. Post-training closes that gap, and understanding why it has become an electrical question requires a short account of how the method works and why the field adopted it.

Reinforcement learning is the branch of machine learning in which an agent improves a decision policy through interaction: the policy chooses actions, the environment returns rewards, and the policy is updated to make higher-reward actions more probable. Policy-gradient methods perform this update by direct ascent on expected reward, and Proximal Policy Optimization, which stabilizes the ascent by clipping how far any single update can move the policy, has become the workhorse of the family because it is simple, robust, and tolerant of imperfect reward signals [21]. Language-model alignment adopted this machinery in the form of reinforcement learning from human feedback: a reward model is trained on human preference comparisons, and the language model is then optimized against it with PPO, which is how instruction-following assistants acquired their behavior. The approach works, but the classical pipeline is heavy, since PPO for language models requires a separate value network as large as the model being trained, and the reward-model stage adds cost and complexity of its own. Direct Preference Optimization simplified the pipeline by removing explicit reinforcement learning [28], and Group Relative Policy Optimization completed the arc in the opposite direction: it keeps reinforcement learning but removes the value network entirely [29]. GRPO's insight is that a baseline for judging an answer does not need to be learned; it can be computed on the spot by comparing each answer against other answers to the same question. The method proved dramatically effective for reasoning tasks and is the training engine behind recent reasoning models [30], which is why it is the workload this paper instruments: it is what the industry's post-training compute increasingly runs.

A GRPO training step consists of two phases. In the rollout phase the model generates, for each prompt in a batch of B prompts, a group of G candidate answers, where G is called the group size. In the update phase the candidates are scored by a reward function, an advantage is computed for each candidate by comparison against the mean of its own group, and the policy is updated on those advantages. Because the advantage is relative within a group, the parameter G plays a double role: it determines the generation compute expended per prompt, and it determines the statistical quality of the learning signal, since the within-group reward variance that carries that signal degenerates as G shrinks.

Two properties of this structure make the workload a candidate for power control. The first is that the rollout phase dominates the energy budget. In our 72B measurements, generation accounts for 99.9% of wall time, and in our 7B measurements the generation phase draws 253 W on average against an idle draw of 86 W, on an accelerator whose nameplate rating is 400 W [31]. The second is that the dominant phase is parameterized at runtime. Group size and batch size are legitimate configuration choices over a broad range [29, 28], and the training framework used in this work reads its generation count freshly at every step, which makes adjustment during a run mechanically possible [32].

One further piece of background is required for the multi-GPU results. A GPU's nameplate rating is a thermal ceiling rather than an operating point; the A100's 400 W rating bounded our observations, with brief excursions to 434 W, while sustained draw varied between 86 W at idle and roughly 250 to 315 W during generation depending on configuration. Models too large for a single accelerator's memory are sharded: a 72B-parameter model is split layer-wise across four accelerators, which then execute as a pipeline in which each device computes its layers and passes activations to the next. A consequence, measured in Section 6.3, is that the devices largely take turns rather than working simultaneously. Their power draws are slightly anti-correlated, and cluster-level power spikes arise only in the brief moments when pipeline stages overlap.

# 4. Methodology

## *4.1 Problem formulation*

We formulate power control of a training job as a Markov decision process solved by a learned policy that observes recent power telemetry and adjusts the training loop's generation parameters. The state is an eight-dimensional vector comprising the five most recent power readings normalized by the applicable power cap, the current headroom defined as the cap minus the latest reading and normalized by the cap, the current training phase encoded numerically, and the current group size normalized to its admissible range. The normalization is a deliberate design decision rather than a convenience: expressing all power quantities relative to the cap is what later allows a single policy to transfer across hardware scales whose absolute power levels differ by hundreds of watts. The action space contains five discrete operations: hold the current configuration, decrement or increment the group size G within its admissible range, and decrement or increment the batch size within the range one to eight. During controller training, transition dynamics are supplied by trace replay. The environment steps through a measured telemetry series, and the agent's group-size setting modulates the replayed power through a multiplicative actuator model in which power equals the trace value scaled by 0.8 plus 0.2 times G over its maximum, a form reflecting the measured dependence of generation load on group size at single-GPU scale. Episodes run 500 environment steps from randomized offsets into the trace.

### *4.2 Controller design and its evolution*

The controller is a Proximal Policy Optimization agent with a multilayer-perceptron actor-critic architecture. Its reward function evolved through four designs, each correcting a measured failure of its predecessor, and we report the sequence in full because two of the failures constitute findings in their own right.

The first design, which we denote G1, carried raw power values in watts in its state and used a reward of plus one per non-violating step, a fixed penalty of minus eight per violation, and a throughput bonus proportional to the product of group size and batch size. Trained on the 7B trace, G1 performed well in its home environment and transferred to nothing else. A policy that has learned that four hundred watts is dangerous has learned nothing useful about a four-GPU cluster whose ordinary draw is 588 W. The second design, G2, replaced raw watts with cap-relative quantities throughout the state, and its training sampled episodes jointly from the 7B, 14B, and 72B trace environments, a domain-randomization strategy. G2 produced a single policy that transfers across scales without retraining, and it is the base policy for all subsequent designs.

The third design, G3, addressed a standing objection to applied reinforcement learning: hand-tuned penalty weights. Our own attempts to tune the violation penalty oscillated between over-enforcement, which collapsed throughput, and under-enforcement, which ignored the cap. G3 therefore replaces the fixed penalty with a Lagrangian formulation drawn from the constrained reinforcement-learning literature [24]. The objective becomes throughput maximization subject to a violation-rate constraint, and the multiplier lambda is updated by dual ascent over a sliding window of 4,000 environment steps, with a step size of 0.5, a ceiling of 20, and the update proportional to the gap between the observed violation rate and the target. Two procedural details proved essential to making this textbook mechanism work in practice. The constraint target must be feasible: before training we measured the attainable violation band on each trace, that is, the violation rates achieved at the minimum and maximum group sizes, and set the target of 8% inside the band for all three environments. An earlier target of 10% lay outside the feasible band of the 14B environment and drove the multiplier to divergence. With a feasible target in place, the multiplier behaved exactly as the theory intends, rising from its initialization of 3 to approximately 12 and stabilizing precisely as the observed violation rate converged onto the target. No penalty weight in G3 was set by hand.

The fourth design, G4, protects the training process from its own controller. Reducing the group size saves power but attacks the statistical signal on which GRPO learning depends: at a group size of two, our training logs record the fraction of groups carrying zero reward variance reaching 0.94, which is near-total signal collapse. Two intuitive protection mechanisms failed in instructive ways. Rewarding signal quality caused the policy to maximize the group size and ignore power entirely, and penalizing signal collapse produced the same behavior through the mirrored gradient; in both cases the signal term dominated the learning landscape. What worked was structural rather than reward-based: a hard floor in the action space, a minimum group size of three set from the measured collapse threshold, inside which the policy optimizes power and throughput freely. The lesson, that signal protection belongs in the action space rather than in the reward, mirrors independently the design of Autodata, which protects data learnability through a hard accept-or-reject criterion rather than a soft quality reward [26]; Section 6.8 reports the full study and the comparison.

All controllers were trained with Stable-Baselines3 [23] over Gymnasium environments [22] using a learning rate of $3\times10^{-4}$, a rollout horizon of 512 steps, minibatches of 128, ten optimization epochs per update, a discount factor of 0.99, generalized advantage estimation with lambda 0.95, a clipping range of 0.2, four parallel environments, and 100,000 to 150,000 total timesteps per controller. Controller training completes in roughly ten minutes on a laptop CPU; no GPU is required.

### 4.3 Live integration

Live operation works through a callback registered with the trainer. At each step boundary the callback reads cluster power through the telemetry interface, assembles the normalized state, queries the policy, and writes the resulting group size directly to the trainer's live attribute, which the framework reads freshly at each step [32]. Two implementation constraints matter in practice. The framework copies its launch configuration once at initialization, so writes to the configuration object are silently ineffective and the live attribute must be set directly; discovering this cost one instrumented hardware run whose controller decisions were logged but never applied. Additionally the generation batch must remain divisible by the group size, so the callback snaps the policy's proposal to the nearest valid divisor. Live operation was subsequently verified on hardware: in the instrumented 72B runs of Section 6.3 the controller's group size moved through sequences such as five, six, five, four, three, two in response to measured power.

### 4.4 Instrumentation and evaluation protocol

All power measurements use the NVIDIA Management Library [33] sampled at 500 ms across every GPU in a job. Each sample records the timestamp, elapsed time, GPU index, training phase, power in watts, GPU and memory utilization, temperature, a cap-violation flag, and a throttle-risk flag; multi-GPU runs additionally record the cluster-aggregate power. Phase labels distinguishing idle, rollout, and update are produced by a callback hooked into the trainer's step lifecycle, so that every power sample is attributed to the activity that produced it, and a token counter attached to the same callback accumulates generated-token counts from the trainer's logs, enabling energy-normalized output metrics.

Controller evaluation replays a full measured telemetry trace through the environment twice under identical conditions, once under a random policy serving as the baseline arm and once under the trained policy acting deterministically. Both arms run 500-step episodes. Violation counts, token totals, energy, cost, and carbon are computed from the resulting power series by integrating power over the sampling interval, with cost evaluated at \$0.08/kWh and carbon at the New York grid intensity of 0.25 kg $CO_2$/kWh. Three hygiene rules, each adopted after an early artifact, apply throughout. Comparisons are made in violation rates rather than raw counts, because arms can differ in sample count. Both arms of any comparison are evaluated at the same power cap, with baselines recomputed from raw telemetry whenever the cap changes; an early cap mismatch produced a spurious result of minus 212% before this rule was instituted. Simulation comparisons are repeated over three random seeds. Two comparison baselines accompany the learned controller in every simulation study: the random policy, and a five-line threshold heuristic that reduces the group size when power exceeds 90% of the cap and raises it below 70%. A third natural baseline, the accelerator's own firmware power cap, requires host-level administrative access that cloud instances do not grant; it is addressed analytically in Section 9 and experimentally in the pilot of Section 10.

### 4.5 Occupancy-controller methodology

The rebuilt controller of Sections 6.4 through 6.6 is specified here for reproducibility. Let P(t) denote cluster power sampled at $\Delta t = 0.5$ s, B the power budget in watts, and $\bar{P}_w(t)$ the rolling mean of P over the trailing window of w seconds:

$$\bar{P}_w(t) = \frac{\Delta t}{w} \cdot \sum_{\tau = t-w}^{t} P(\tau) \tag{1}$$

The budget is defined on the 30-second window, and the violation rate over N samples is

$$v = \frac{1}{N} \cdot \sum_t \mathbb{1}[\bar{P}_{30}(t) > B] \tag{2}$$

with the integrated excess in watt-seconds

$$X = \sum_t max(0, \bar{P}_{30}(t) - B) \cdot \Delta t \tag{3}$$

The control problem is constrained throughput maximization: the agent chooses the generation batch level b_k ∈ {1, 2, 4, 8} at each batch boundary k to solve

$$max_\pi \ \mathbb{E}_\pi[\sum_k T_k] \quad subject\ to \quad v \leq \delta,\ \delta = 1\% \tag{4}$$

where T_k is the actual non-padding token count of batch k. The state at each boundary is the eight-vector

$$s_k = [P(t)/B,\ \bar{P}_{10}/B,\ \bar{P}_{30}/B,\ (B - \bar{P}_{30})/B,\ (P(t) - P(t-5))/B,\ b_k/8,\ b_{k-1}/8,\ d_{k-1}/20] \tag{5}$$

with d the batch duration in seconds, and the per-batch reward implements the Lagrangian relaxation of (4):

$$r_k = \frac{T}{1000} - \lambda X_k - 2\lambda\, max(0, \bar{P}_{30,k}^{peak} - B) \tag{6}$$

where X_k is the excess (3) accrued during batch k. The multiplier is not hand-tuned; it follows dual ascent toward the violation target after every eight episodes with observed rate $\hat{v}$:

$$\lambda \leftarrow clip(\lambda \cdot exp(\eta \frac{\hat{v} - \delta}{\delta}),\ 10^{-3},\ 2),\ \eta = 0.5 \tag{7}$$

The training environment replays the calibration run's measured transition traces: each of the 75 generation batches contributes its raw half-second power segment, token count T, and duration d, banked by level; an episode replays 600 simulated seconds, warm-started at the batch-4 steady state of 462 W. Transitions are split 70/30 into training and held-out validation banks (stratified, seed 1234). PPO hyperparameters: learning rate $3 \cdot 10^{-4}$, 256 steps per update, batch 128, 10 epochs, $\gamma = 0.99$, GAE $\lambda = 0.95$, clip 0.2, entropy coefficient 0.01 to 0.02, for 45,000 to 100,000 environment steps per seed; three seeds were trained and the single seed passing the fixed validation criterion was retained, with all three reported in Section 6.5. The hysteresis baseline drops one level when $\bar{P}_{30} > 512$ W and climbs when $\bar{P}_{30} < 510$ W and $\bar{P}_{10} < 525$ W, applied unchanged in absolute watts at the 561 W hardware budget. The hardware budget follows a normalization rule fixed before the comparative runs,

$$B_{new} = B_{cal} \cdot \frac{\bar{P}_{4,new}}{\bar{P}_{4,cal}} \tag{8}$$

where $\bar{P}_4$ is the plant's measured batch-4 mean power; the rule preserved the batch-4-to-budget ratio of 0.888 across a 44 W plant shift. Arms ran in randomized order across three replications on two physical pods.

## 5. Experimental preparation

The experiments comprise two campaigns. The first is a controlled progression across scale, holding the model family, dataset, training method, and instrumentation constant while varying only the parameter count and the number of accelerators: BF16 GRPO training of the open Qwen2.5 instruction-tuned family [34] at 7B, 14B, and 72B parameters as implemented in the TRL framework [32]. The second campaign, motivated by the first campaign's multi-GPU nulls, runs on the 72B model in AWQ-quantized generation-only form: the actuator-authority sweep, a randomized occupancy-calibration run, smoke and thermal-check runs, and twelve final controller arms executed on two physical pods, including one discarded thermally misconfigured warm-up run that is disclosed in the artifact record. Prompts come from the UltraFeedback binarized preference dataset, whose records supply a prompt together with chosen and rejected responses. The instrumentation runs use a simple length-target reward that scores completions by proximity to a 128-token target. This choice is deliberate: the purpose of these runs is to drive realistic generation load under a controlled reward, which is what power measurement requires, not to advance model quality. The launch configuration sets the group size to four and the per-device batch to four, with bf16 precision throughout the BF16 GRPO campaign.

Table 1 lists every hardware run with its purpose, outcome, and cost. The 7B baseline run of 500 steps produced 40,624 telemetry samples and serves as the primary trace for controller training and evaluation; a second 7B run of 50 steps produced the cleanly phase-labeled telemetry used for the phase analysis; the 14B campaign produced a duration-matched 50-step pair (4,168 and 4,148 samples), and the 72B campaign produced baseline and controller arms of 200 steps each. Across both campaigns the instrument recorded over 380,000 half-second per-GPU samples, and the entire experimental program, including model transfers, smoke tests, and failed attempts, cost roughly $260 of compute.

***Table 1. Hardware run inventory***

| Run | Model / hardware | Steps | Purpose and finding |
|---|---|---|---|
| 1 | 7B / 1×A100 | 500 | Primary baseline; 43.2% violation rate at 314.6 W cap; source trace for controller training and evaluation |
| 2 | 7B / 1×A100 | 50 | Phase-labeled telemetry separating rollout from update |
| 3 | 14B / 2×A100 | 50 + 50 | Duration-matched two-accelerator pair; first null result |
| 4 | 72B / 4×A100 | 200 + 200 | Four-accelerator baseline and controller arms; second null |
| 5 | 72B / 4×A100 | 200 | Live actuation verified; cap-matched comparison; third null |
| 6 | 72B / 4×A100 | 200 | Signal-safe controller deployed live |

| Run | Model / hardware | Steps | Purpose and finding |
|---|---|---|---|
| 7 | 72B AWQ / 4×A100 | 16 sweep segments | Actuator-authority sweep: concurrency 17 to 22% swing, firmware denied |
| 8 | 72B AWQ / 4×A100 | 75 transitions | Randomized occupancy calibration; transition traces for controller training |
| 9 | 72B AWQ / 1-4×A100 | smoke + thermal checks | Load, permission, budget-normalization verification; one discarded warm-up |
| 10 | 72B AWQ / 4×A100, two pods | 12 arms × 20 min | Final four-arm comparison, three replications, randomized order |

The hardware is commodity cloud GPU infrastructure built on NVIDIA A100-SXM4-80GB accelerators. Three deliberate choices deserve justification because each shapes what the measurements mean. The first is the device itself. The A100 was selected because it is the most widely deployed datacenter training accelerator of its generation, its 400 W thermal class is representative of the fleet installed in AI facilities today, and its power telemetry interface is mature and well documented [31, 33]; results measured on it therefore transfer to the installed base rather than to an exotic corner of the market. Table 2 situates the choice within the accelerator landscape that AI datacenters currently provision for, which industry analyses report is dominated by NVIDIA parts [3, 35]. The second choice is the accelerator count, which was not free: it is dictated by model memory. A 7B model in bf16 precision fits comfortably on one 80 GB device, a 14B model requires two, and a 72B model requires four, so the scale progression of the study is simultaneously a sharding progression, from a monolithic single-device job to a two-way and then four-way pipeline. That coupling is not a confound but the object of study, since sharding is precisely the mechanism through which Section 6 finds control authority is lost. The third choice is measuring GPU power rather than whole-node or CPU power. Training large language models is dense matrix arithmetic at extreme parallelism, work for which GPUs deliver an order of magnitude or more throughput per watt than CPUs; in an AI training node the accelerators dominate the electrical load while the host CPU's share is small and approximately constant, and prior characterization of LLM infrastructure accordingly treats the GPU as the unit of power analysis [12]. The controllable structure this paper studies also lives entirely on the accelerator, since the generation workload that the group size modulates executes there. The accelerator count is therefore matched to model scale: one for 7B, two for 14B, and four for 72B.

***Table 2. Datacenter AI training accelerators and their power class***

| Accelerator | Vendor | Nameplate (TDP) | Note |
|---|---|---|---|
| A100 SXM 80GB | NVIDIA | 400 W | This study; largest installed base of its generation |
| H100 SXM | NVIDIA | 700 W | Current mainstream deployment |

| Accelerator | Vendor | Nameplate (TDP) | Note |
|---|---|---|---|
| B200 | NVIDIA | ~1,000 W | Ramping; raises the per-device stakes of provisioning |
| MI300X | AMD | 750 W | Principal non-NVIDIA training part |

Rising nameplate ratings across accelerator generations make the provisioning question of this paper more consequential with each hardware cycle, since capacity purchased against sticker ratings scales with those ratings while measured demand, as Section 6 shows, does not follow them proportionally. The 72B model is loaded across its four devices with an explicitly precomputed device map, after the framework's automatic placement exhibited pathological quadratic scanning at this scale that degraded load times from minutes to hours.

## 6. Results

The results are organized as the experiments were: a progression across scale. The single-accelerator study establishes that the workload has controllable power structure and that the learned controller exploits it. The two-accelerator study produces the first null result. The four-accelerator study replicates the null under verified live actuation and yields the mechanism. A cross-scale comparison then separates what learning contributes from what any rule could achieve.

### *6.1 Single-accelerator study: 7B on one A100*

The 500-step 7B run characterizes the workload's power anatomy. Generation draws a sustained 250 to 315 W with the run's mean at 299.2 W and brief peaks to 410.6 W, against an idle draw of 86 W, and the phase-labeled run shows the power series rising and falling with the rollout-update cadence of the training loop. Figure 1 displays a segment of the phase-labeled trace. Against the evaluation cap of 314.6 W, the baseline run spends 43.2% of its samples in violation, a total of 216 violating samples over the 500 steps. This is a sustained-overload condition rather than an occasional excursion; it emulates the situation of commercial interest, in which a facility oversubscribes its capacity aggressively or a demand-response event requires immediate load reduction, and Section 7 will show that this sustained class of violation is precisely the class that survives every measurement window and therefore matters to infrastructure.

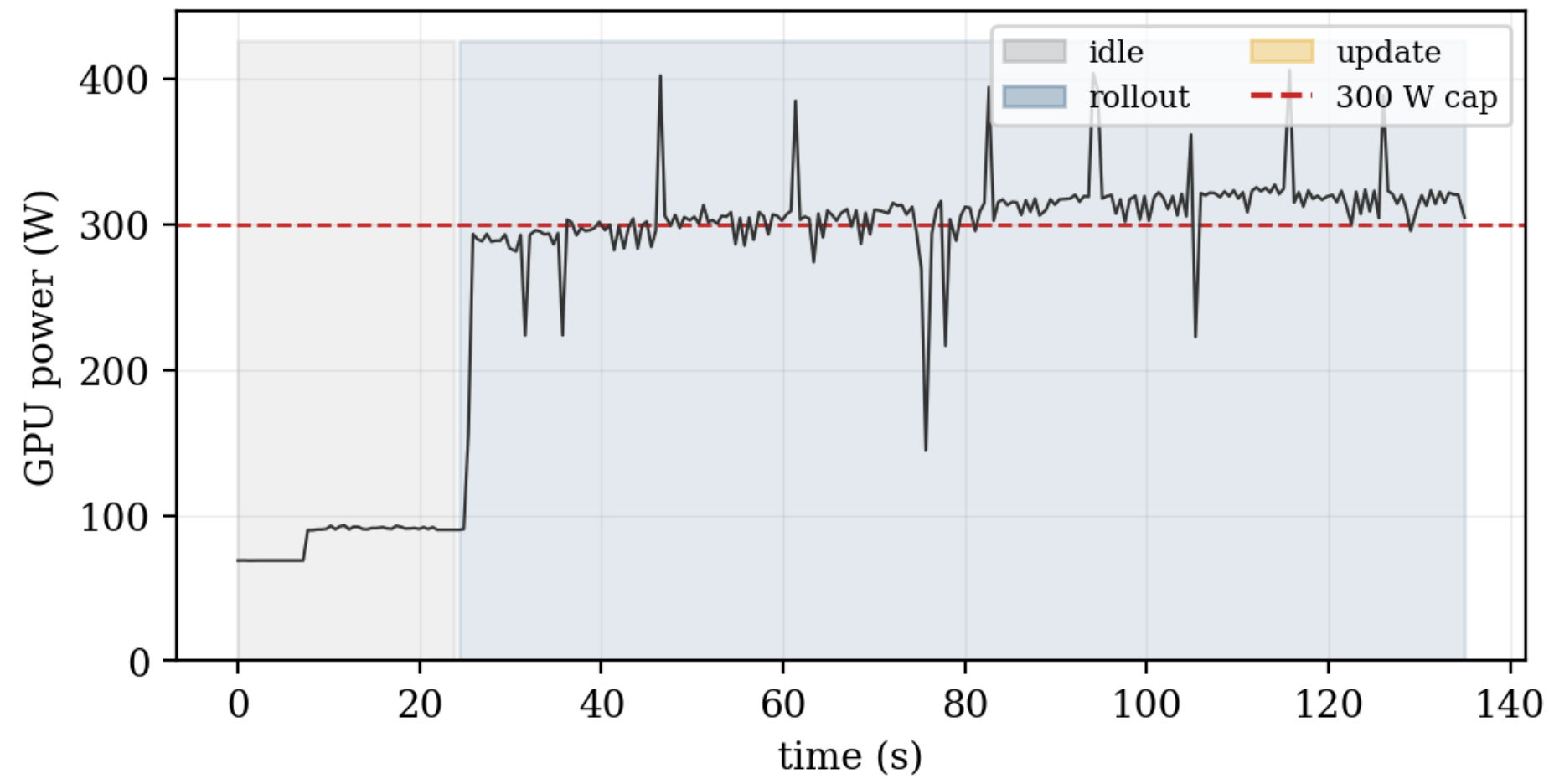

*FIGURE 1. GPU power during GRPO training with phases labeled, 7B model on one A100. The rollout phase carries the sustained load; the sticker rating of 400 W is touched only in brief excursions.*

Evaluating the trained controller against this full measured trace under the protocol of Section 4.4 yields the results of Table 3. Violations fall from 216 to 22, a reduction of 89.8%, while token output rises 18.1% and energy consumed falls 6.5%, so that energy efficiency improves by 26.2% in tokens per megawatt-hour. Mean generation-phase power falls by 19.3 W and peak power by 35.8 W. The simultaneous improvement of the power and output metrics is not paradoxical; it is the signature of selective control. The policy reduces the group size only in the moments the power history marks as critical and recovers throughput through the batch dimension when headroom returns, in contrast to a firmware power cap, which throttles clock frequency indiscriminately across both phases of the loop regardless of the moment's criticality. Figure 2 overlays the baseline and controlled power series against the cap.

***Table 3. Controller evaluation on the measured 7B trace (500 steps, 314.6 W cap)***

| Metric | Baseline | Controller | Change |
|---|---|---|---|
| Cap violations | 216 | 22 | −89.8% |
| Violation rate | 43.2% | 4.4% | −38.8 pp |
| Total tokens | 1,731,712 | 2,044,928 | +18.1% |
| Mean power (W) | 299.2 | 279.9 | −19.3 W |
| Peak power (W) | 410.6 | 374.7 | −35.8 W |
| Energy, full run (kWh) | 1.777 | 1.662 | −6.5% |
| Tokens per MWh | 0.975 B | 1.230 B | +26.2% |

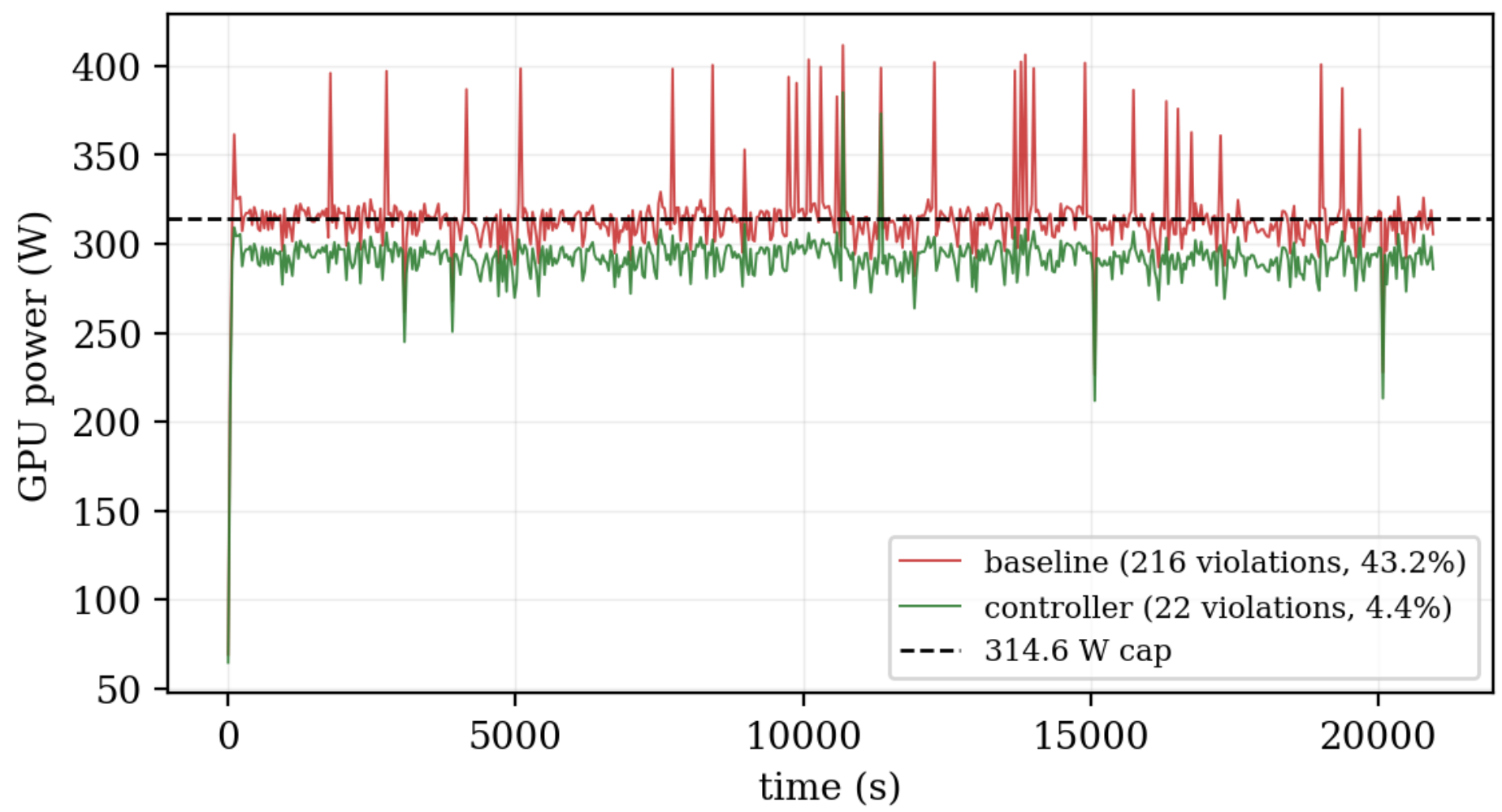

*FIGURE 2. Baseline and controller power series over the full 500-step 7B trace with the 314.6 W cap drawn. The controller removes the violation mass while the trace remains active.*

### *6.2 Two-accelerator study: 14B on two A100s*

The 14B study repeats the design at doubled scale with a duration-matched pair: a 50-step baseline arm and a 50-step controller arm on two accelerators, with the model sharded across both, running 1,136.6 and 1,128.3 seconds respectively. The matching matters because it removes every confound a critic could reach for: the two arms consumed identical energy, 0.110 kWh each, at identical mean per-device power, 190.7 against 191.1 W. The baseline violation rate at the 300 W per-device cap is 7.99%, an order of magnitude sparser than the 7B condition, because sharding halves each device's share of the sustained load. Against this baseline, the controller arm records 8.10%, a change of minus 0.9%, indistinguishable from zero. At the time, three explanations seemed available: that the controller, trained on single-GPU data, did not recognize two-GPU power patterns; that the per-device cap left too little violation mass to reduce; or that something deeper was wrong with the actuator itself. The single-scale retraining and cap adjustments motivated by the first two explanations did not move the result, which pointed to the third and motivated the instrumented campaign at the next scale.

### *6.3 Four-accelerator study: 72B on four A100s*

The 72B study is the decisive one, comprising a 200-step baseline arm and three successive live controller arms with different policies: the transferred multi-scale policy, a cluster-trained variant whose state carries the aggregate rather than per-device power, and the signal-safe design. The baseline characterizes the cluster's power behavior: per-device draw averages 147.0 W, cluster draw averages 588 W with excursions to 1,221 W, and the per-device violation rate at the evaluation cap is 25.92% when recomputed under the cap-matching rule of Section 4.4. Against this baseline the three controllers change the violation rate by +4.4%, −1.6%, and −2.7% respectively: three essentially null results from three different policies. Critically, the third and fourth runs verify that the null is not an integration failure. The controller's decisions were logged applying to the trainer's live attribute, and its group size moved through sequences such as five, six, five, four, three, two in response to measured power. The agent acted; the power did not move.

Three measured facts jointly explain the boundary, and Figure 3 visualizes the first of them. First, the actuator loses authority under sharding: mean per-device power is 147.0 W with the controller and 147.0

W without it, identical to the tenth of a watt, because a pipeline of sharded layers executes at its natural rate regardless of how many candidate answers each prompt receives. The group size that dominates single-device load becomes a bystander at cluster scale. Second, the disturbance changes character. The pairwise correlation of the four devices' power draws is −0.13; the accelerators take turns, and the cluster's peaks are the brief coincidences of pipeline stages, with a median duration of 0.5 seconds and a ninetieth percentile of 2 seconds (Figure 4). Third, the control loop is mismatched to that disturbance by two orders of magnitude: the controller acts once per training step, approximately every 55 seconds, which is also the dominant periodicity of the cluster power spectrum, while the disturbances it would need to reject live below two seconds.

The study also quantifies a gap of methodological consequence. The replay environment, whose actuator model was calibrated at single-GPU scale, credited these same controllers with up to 65% violation reduction at the 72B scale; the hardware showed approximately zero. Trace-replay simulators with parametric actuator models systematically overestimate control authority, and hardware validation is mandatory. This finding governs the design of Section 8, whose fleet analysis uses only raw measured traces and never the actuator model.

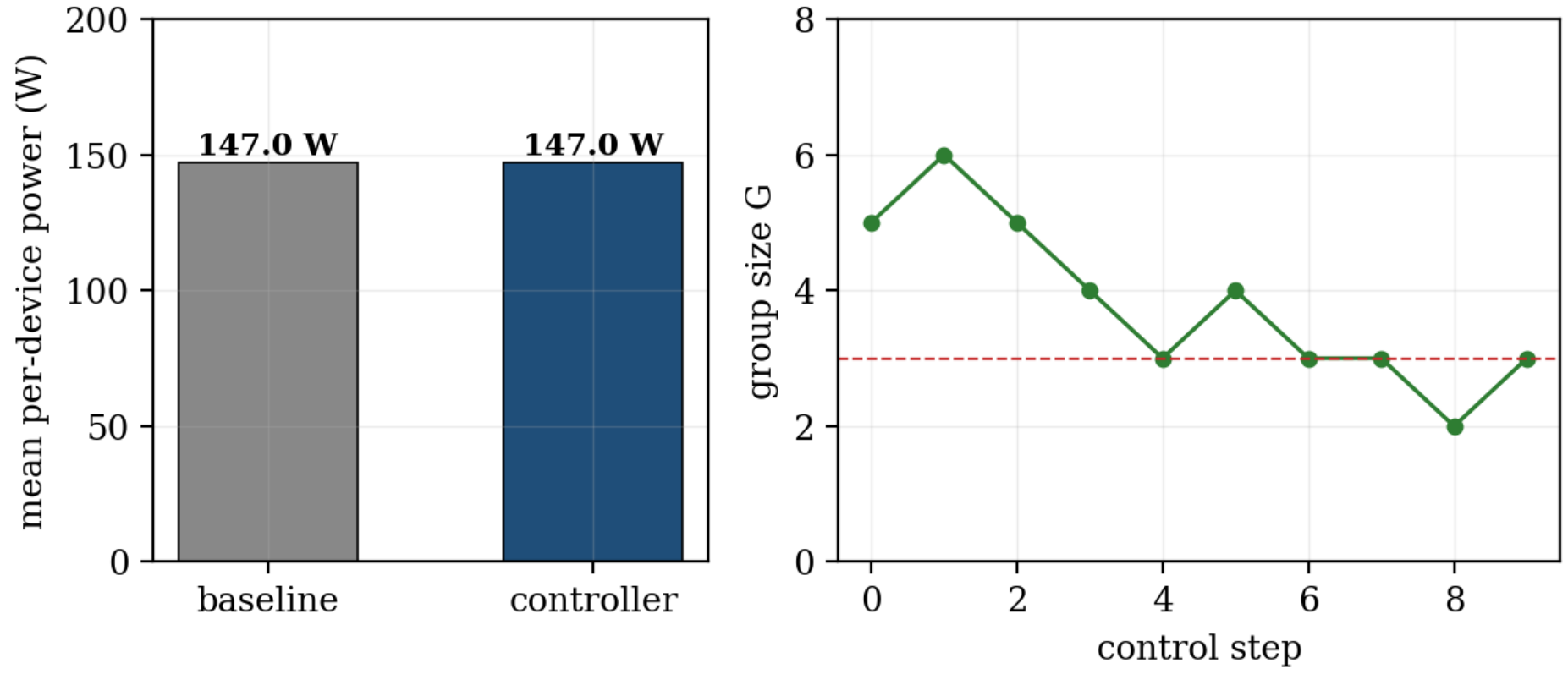

*FIGURE 3. Left: mean per-device power with and without the controller in the 72B live runs, identical at 147.0 W. Right: the controller's group-size trajectory during the same run, demonstrating that the null reflects lost actuator authority rather than absent actuation.*

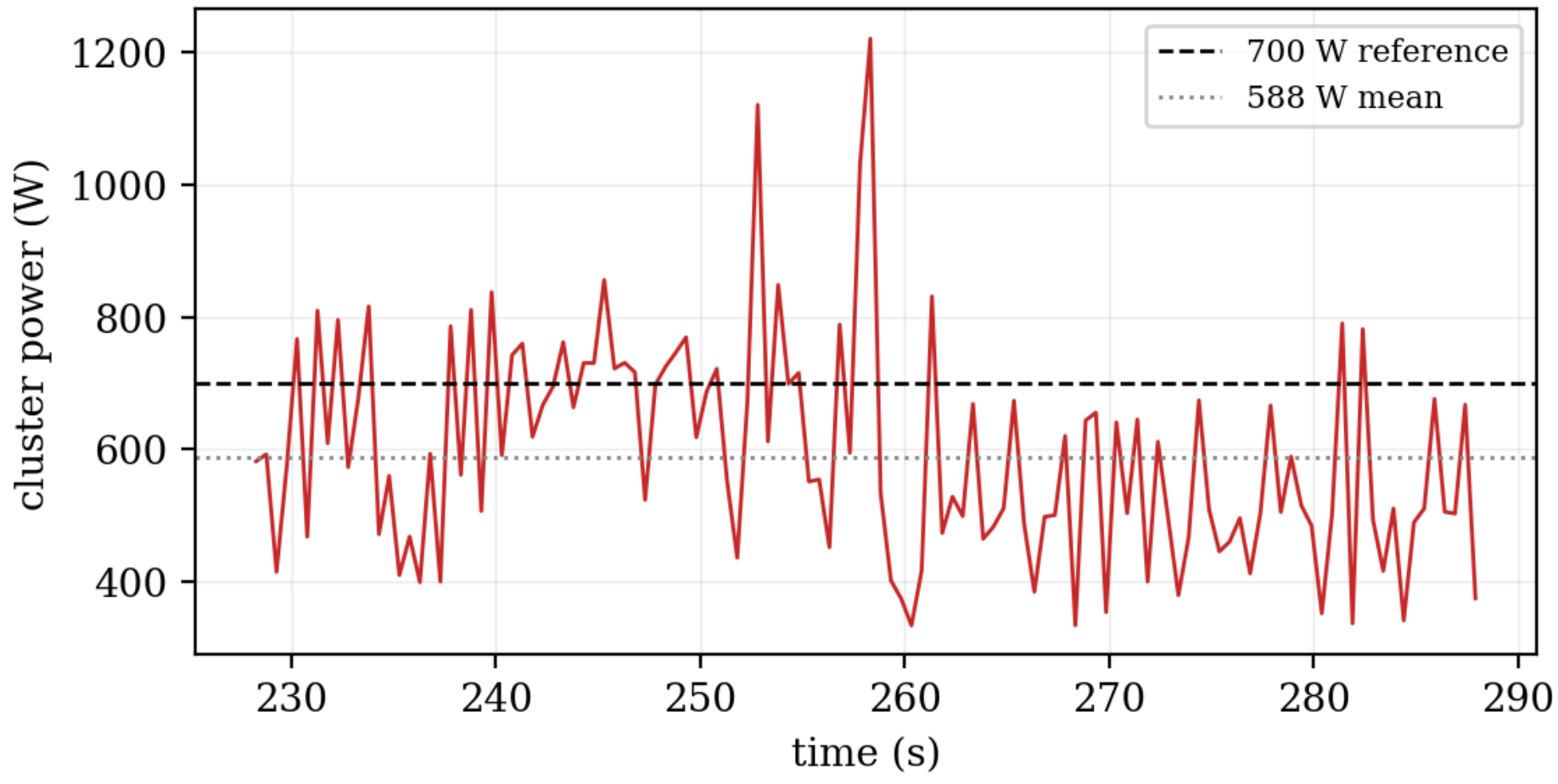

*FIGURE 4. Sixty seconds of 72B cluster power. The spikes above the 588 W mean are momentary pipeline-stage overlaps with median duration 0.5 s.*

### *6.3.1 Beyond group size: the actuator hierarchy at scale*

An obvious objection: having found that group size fails, why not try other knobs until one works? Because the measured disturbance dictates which actuators can possibly work; reasoning from the measurements maps the control space instead of sampling it.

The governing facts are three, all established above: the disturbance at 72B is sub-second, with spikes of median duration 0.5 seconds; group size cannot move sustained per-device power under sharding, since mean draw was identical to the tenth of a watt with and without the controller; and any actuator embedded in the training loop can act no faster than one step, roughly every 55 seconds. These facts partition the knobs. Software parameters read once per training step share a timescale ceiling: acting every 55 seconds, they cannot reject a half-second disturbance regardless of their power authority. Applied at generation-batch boundaries instead, the same parameters escape that ceiling by a factor of three to four (Section 6.4); the sub-second transient itself remains beyond any software loop. Reducing generation to sub-step chunks could in principle tighten the loop, but at the cost of restructuring the trainer and paying synchronization overhead every chunk, and it still acts through the same sharded pipeline whose rate is set by its slowest stage. Actuators matching the disturbance lie outside the training loop. Firmware power limits and locked clocks, set through the management library, act in milliseconds on each device independently of shard structure, and are therefore the only in-node actuator whose response time and authority both suit the sub-second transient; their cost is that they require host-level access unavailable inside our cloud containers, which is why the present study could not exercise them and why they are the first target of the validation pilot. Beyond the node, the fleet scheduler acts on the timescale of job placement, and Section 8 shows this is where multi-GPU jobs are properly governed, since their strong 55-second periodicity and stable means make them predictable inputs to external staggering.

The conclusion is not that reinforcement learning failed at 72B; the control layer must change with the parallelism regime. At single-device scale the workload's power lives in a software parameter the training loop can reach, and the learned controller succeeds. Under sharding that same parameter loses authority, the dominant disturbance moves to a sub-second timescale no training-loop actuator can reach, and control must move with it, downward to firmware and outward to the scheduler. This is a design principle, and Sections 6.4 through 6.6 complete it constructively: an actuator-authority sweep locates a software knob that does retain authority at 72B scale, a controller is rebuilt around it, and live hardware control is demonstrated across three replications. The correct summary of the multi-GPU boundary is therefore that the group-size actuator loses authority under sharding, not software control as such.

### *6.4 The actuator-authority sweep*

The hierarchy of Section 6.3.1 predicts which actuators can work; the next experiment measures which do. Before building any new controller we swept every user-space knob available on the 72B condition, running short fixed-setting generation segments per value while recording cluster power, throughput, and signal statistics. Two disclosures apply. First, the sweep and everything downstream of it use an AWQ-quantized generation-only proxy for the rollout phase rather than the BF16 GRPO training loop of Sections 6.1 through 6.3; the proxy is justified because generation accounts for 99.9% of measured 72B wall time and therefore essentially all of its power, but the two workloads are distinguished throughout and the original null results stand as reported for the full training loop. Second, the firmware actuators, per-device power limits and locked clocks, were probed first and denied by the cloud container with insufficient permissions, converting what would have been the strongest sweep rows into a measured deployment boundary: firmware power control requires operator-level infrastructure access that commodity cloud instances do not grant.

Table 4 and Figure 5 report the sweep. The result inverts the training nulls' expectation. Group size, applied here as generation concurrency, moves mean cluster power by 16.9% and the 30-second peak by 17.0% while throughput scales 3.3 times; the generation batch size, the purest concurrency knob, moves mean power by 22.5% and the peak by 24.0% at 5.2 times throughput. Completion length moves power by only 3.5% and an artificial inter-batch pause by 6.3%, both below the 10% authority threshold. Energy efficiency moves with concurrency as strongly as power does, falling from 2.97 to 0.74 Wh per thousand tokens across the batch range, a fourfold improvement at high occupancy.

***Table 4. 72B actuator-authority sweep (generation proxy, 4 A100s, 20 steps per value)***

| Knob | Mean-power swing | 30 s peak swing | Throughput ratio | Verdict |
|---|---|---|---|---|
| Generation batch size (concurrency) | 22.5% | 24.0% | 5.2x | Has authority |
| Group size (as concurrency) | 16.9% | 17.0% | 3.3x | Has authority |
| Completion length | 3.5% | 1.8% | 1.2x | No authority |
| Inter-batch pause | 6.3% | 3.0% | 1.3x | Marginal |
| Firmware power cap / clocks | denied | denied | denied | Requires operator access |

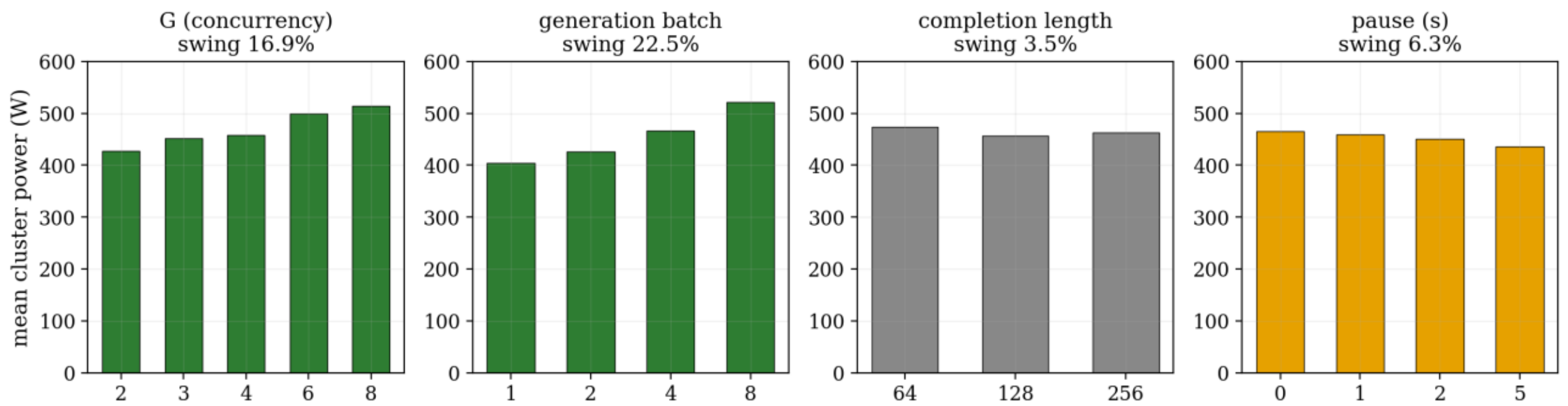

*FIGURE 5. Mean cluster power across the sweep values of each software knob. Concurrency knobs (batch, group size) move 72B power by 17 to 22 percent; completion length and pause do not.*

The sweep resolves the nulls rather than contradicting them. In the training loop, changing the group size changed the volume of generation work, which the sharded pipeline processed at its own rate, leaving per-device power untouched; applied as concurrency, the same parameters change how many sequences occupy the pipeline simultaneously, and occupancy is what draws power. Actuator authority is therefore implementation-dependent: a knob has power authority precisely insofar as it changes pipeline occupancy rather than work volume. This principle explains every 72B result.

### *6.5 Rebuilding the controller on the occupancy actuator*

The controller was rebuilt around the validated actuator. A twenty-minute randomized calibration run on the four-A100 cluster hopped the batch level among {1, 2, 4, 8} at generation boundaries, producing 75 level transitions with raw half-second telemetry, per-batch durations of 12 to 21 seconds, and actual non-padding token counts per batch; these transition traces, not summary statistics, form the training environment. The controller observes an eight-feature state (instantaneous, 10-second, and 30-second budget-normalized cluster power, 30-second headroom, a five-second slope, and the current, previous, and duration context of the batch level) and selects the next generation batch level at each batch boundary, a 12-to-21-second decision cadence that is honest to the blocking nature of generation and still three to four times tighter than the training loop's 55-second step. Training used the constrained formulation of Section 4.2: throughput maximization with the violation penalty's multiplier set by dual ascent against a 1% violation-rate target, three seeds, and retention of the single policy passing a fixed validation criterion on held-out calibration transitions; one of three seeds passed, and seed-level variability is reported rather than hidden.

The simulation study first mapped feasibility; the map is itself a finding. At a 500 W thirty-second budget, no tested PPO policy or cyclic mixing schedule satisfied the pre-specified criterion of at least 10% more tokens than the static batch-4 baseline at 1% violations or fewer: batch 8 draws a mean of 510.5 W, above the budget itself, so sustained time at the high level drags the rolling mean over. The cyclic-schedule search and all three PPO seeds independently converged to the static batch-4 policy, which at that budget is the correct optimum; a controller that correctly identifies inaction as optimal validates the method. Raising the budget to 520 W, above the high level's mean draw, makes the criterion attainable, and the selected controller achieved 94,349 tokens at 0.49% violations on held-out transitions, a 40% gain over static batch-4, while a grid search over hysteresis thresholds found no threshold configuration able to reach 1% violations at all: a boundary-actuated greedy rule that climbs whenever headroom appears books roughly twenty seconds of violations at every crossing before its next decision, whereas the learned policy discovers a duty-cycle that spends a fixed fraction of time at the high level regardless of instantaneous headroom. The value of occupancy control is thus sharply budget-dependent, with a phase boundary at the high level's mean draw: below it control is worthless and the controller correctly abstains; above it, mixing policies unlock 27 to 46% more output at under 1% violations.

### *6.6 Live four-arm hardware study*

Four arms ran live on four A100s at equal twenty-minute wall-clock durations: uncontrolled batch 8, static batch 4, the optimized hysteresis rule, and the selected PPO controller, each replicated three times across two physical pods with arm order randomized. Budget selection followed a normalization rule fixed before the comparative runs: a live smoke test at the simulation's 520 W budget failed with 87% violations because the hardware pods draw roughly 44 W more than the calibration pod at identical settings, so the budget was retargeted to 561 W to hold the batch-4-to-budget ratio at its calibration value, after which a second smoke test recorded 0.0% violations; the cap-normalized state transferred across the plant shift with a one-number budget correction, while the watt-absolute hysteresis thresholds had to be re-derived. One third-replication warm-up run at a mis-set 530 W budget, caused by reading cold-GPU thermals, was discarded and rerun at 561 W; all runs including the discarded one are disclosed in the artifact record.

***Table 5. Live four-arm results (561 W / 30 s budget, 20 min per arm, n = 3, mean and sample standard deviation)***

| Arm | Actual tokens | Tokens/s | Violation rate (30 s) | Integrated excess (W s) | Wh per 1k tokens |
|---|---|---|---|---|---|
| Static batch 8 | 222,472 ± 5,549 | 189.1 ± 4.0 | 17.72% ± 11.40 | 1,934 ± 1,500 | 0.802 ± 0.008 |
| Static batch 4 | 134,860 ± 10,607 | 114.5 ± 9.0 | 0.01% ± 0.02 | 0.03 ± 0.06 | 1.219 ± 0.072 |

| Arm | Actual tokens | Tokens/s | Violation rate (30 s) | Integrated excess (W s) | Wh per 1k tokens |
|---|---|---|---|---|---|
| Optimized hysteresis | 166,189 ± 9,106 | 141.1 ± 8.3 | 1.96% ± 0.57 | 132 ± 73 | 1.019 ± 0.056 |
| PPO occupancy controller | 183,069 ± 9,093 | 154.8 ± 7.3 | 2.27% ± 1.08 | 159 ± 8 | 0.947 ± 0.029 |

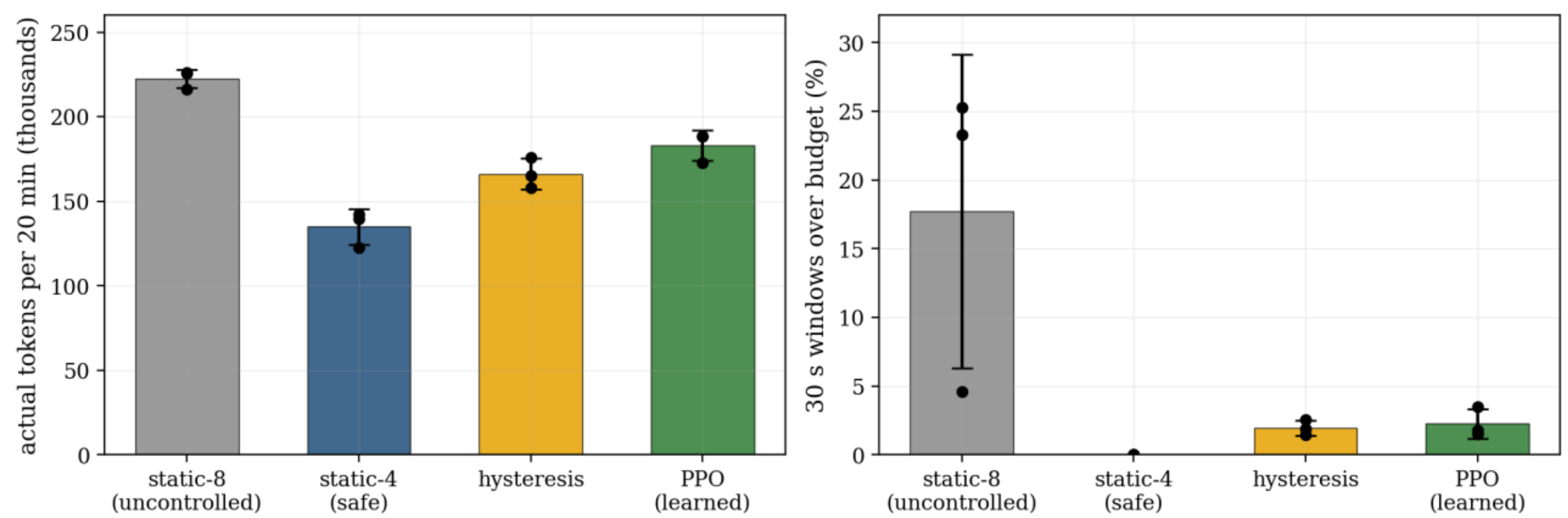

*FIGURE 6. The live four-arm study at 72B: actual tokens per twenty-minute arm and 30-second violation rate, mean and standard deviation over three replications on two physical pods.*

Against the pre-specified criteria: positive, with one miss. The learned controller produced 35.7% more tokens than the static safe baseline, against a required 10%, and 22.3% lower energy per token, while its violation rate of 2.27 ± 1.08% exceeded the strict 1% bar; the per-replication rates were 3.51, 1.78, and 1.52%, run in randomized order across differing operating conditions, and the integrated exceedance of 159 ± 8 watt-seconds was nearly constant across pods and replications, a 5% relative spread against the heuristic's 55%, which is the property an operator prices. Relative to the uncontrolled arm, the controller retained 81.9% of the throughput while cutting the violation rate by 87.2% and the integrated exceedance by 91.8%. Against the optimized heuristic the comparison is favorable but not uniform: the controller delivered 10.2% more tokens and 7.1% lower energy per token on average, winning the paired comparison in the first two replications, while in the third replication, on a pod whose operating point had drifted roughly 11 W cooler, the heuristic adapted more aggressively to the easier environment and marginally outperformed the fixed learned duty-cycle. The accurate claim: the learned controller achieved the highest mean throughput and the lowest mean energy per token among the constrained controllers, while the heuristic had a lower mean violation rate and won the third replication; per-replication values are reported alongside means.

## 6.7 Cross-scale comparison of policies

Table 6 places all controller generations against the random and heuristic baselines in matched three-seed simulation on the real traces of all three scales, using the evaluation protocol of Section 4.4. Three observations organize the table. The learned policies dominate the threshold heuristic on violation reduction wherever violations are dense, by roughly a factor of four at identical sensors and actuator, which isolates the contribution of learning itself. The constrained controller exhibits the economically correct behavior that no static rule reproduces: where its constraint is slack, as in the naturally sparse 14B environment, it converts the slack into a 34% throughput gain, and where the constraint binds it enforces to the target, landing at 7.2% against the 8% target in the 72B environment. The signal-safe design achieves the best joint operating point at scale, a 65.3% violation reduction at a 4.2% token gain with the group-size floor never breached, while its two failed predecessors establish the structural-constraint finding of Section 4.2.

***Table 6. Policy comparison in simulation on real traces (violation reduction / token change vs random; 3 seeds)***

| Policy | 7B (300 W) | 14B (300 W) | 72B cluster (700 W) |
|---|---|---|---|
| Threshold heuristic | +5.9% / −24.0% | 0% / +44.9% | +11.5% / +2.2% |
| G2: normalized PPO | +46.1% / −53.1% | 0% / +179.8% | +42.3% / −13.7% |
| G3: constrained PPO | +30.3% / −11.8% | −2.4% / +34.2% | +57.4% / −24.3% |
| G4: signal-safe PPO | +26.9% / −41.0% | +6.7% / +25.3% | +65.3% / +4.2% |

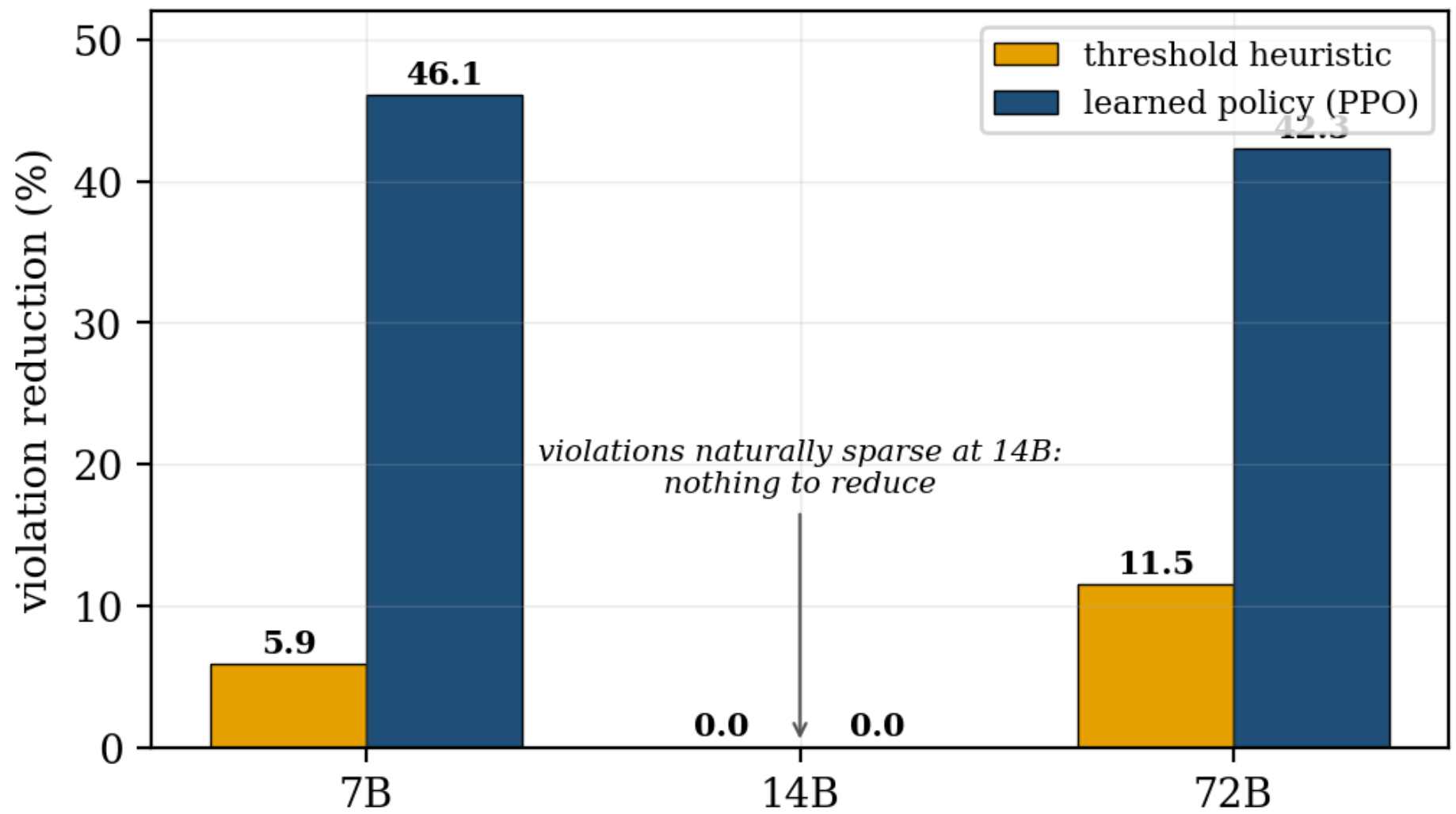

*FIGURE 7. Violation reduction by policy and scale, isolating the contribution of learning over the threshold rule at identical sensors and actuator.*

## *6.8 The signal-protection study and the comparison with Autodata*

Two internal studies, both conducted in the simulation environment on the real traces, deserve their own account because they carry the paper's principal methodological findings. The first concerns the constrained controller. Before training G3 we measured the feasible violation band of each trace environment, the rates attainable at the minimum and maximum group sizes, and this measurement proved decisive: a constraint target of 10%, set before the band was known, lay outside the feasible region of the 14B environment and drove the Lagrange multiplier to divergence, while the 8% target chosen inside the band produced textbook behavior, with the multiplier rising from 3 to approximately 12 and stabilizing exactly as the observed violation rate converged onto the target. The procedural lesson is that constrained reinforcement learning requires feasibility measurement before target selection, a step the literature rarely makes explicit.

The second study is the signal-protection series, and it is where this work meets Meta's Autodata most directly. Reducing the group size saves power but erodes the statistical signal GRPO learns from: our training logs record the fraction of groups with zero reward variance reaching 0.94 at a group size of two, which is a collapsed gradient. We built three successive protections, summarized in Table 7. The first version added a reward bonus proportional to signal quality; the trained policy maximized the bonus by holding the group size at its ceiling and ignored power entirely. The second version replaced the bonus with a penalty on signal collapse; the policy found the mirrored strategy through the same gradient and again abandoned the power objective. The third version moved the protection out of the reward and into the action space, a hard floor at a group size of three set from the measured collapse threshold, inside which the policy optimizes power and throughput freely; it achieved a 65.3% violation reduction at a 4.2% token gain in the 72B environment with the floor never breached.

***Table 7. The signal-protection series (simulation on real traces, 3 seeds, 72B environment)***

| Version | Protection mechanism | Outcome |
|---|---|---|
| v1 | Reward bonus for signal quality | Policy maximizes G, ignores power entirely |
| v2 | Reward penalty on signal collapse | Mirrored failure through the same gradient |
| v3 | Structural action-space floor ($G \geq 3$, from measured collapse threshold) | +65.3% violation reduction, +4.2% tokens, floor never breached |

The comparison with Autodata [26] sharpens both what is shared and what is new. Autodata is an agentic system that generates and selects synthetic training data under a compute budget, and its central safeguard is learnability protection: candidate data is admitted only if a measured gap between weak and strong solvers certifies that the data carries trainable signal, a hard accept-or-reject criterion rather than a soft score. The present work optimizes a different resource, electrical power, under a different budget, facility capacity, and protects a different signal, the within-group reward variance on which GRPO learning depends; yet both systems arrived independently at the same design conclusion, that the learning signal must be protected by a hard constraint and not by a reward term, and our v1 and v2 failures constitute an experimental demonstration of why the soft alternative fails. The novelty of this work relative to Autodata is the axis of operation: Autodata acts on the data pipeline before and between training runs, while our controller acts on the infrastructure coupling of the training loop itself, in real time, on live hardware, with a reinforcement-learning agent governing another reinforcement-learning process. To our knowledge neither the workload characterization nor this meta-control configuration has been reported before.

### *6.9 Energy, cost, and carbon accounting*

Because every power sample carries a timestamp and every run carries a token count, the telemetry supports a complete energy ledger, and the ledger is where the controller's value becomes legible in operator units rather than in violations. Table 8 reports it for the 7B evaluation. Both arms are integrated over the same complete 20,312-second telemetry timeline used for the token counts, 40,624 half-second samples whose full-timeline mean of 314.9 W sits above the 299.2 W generation-phase mean of Table 3 because it includes inter-step and update phases. On that timeline the baseline consumes 1.777 kWh and the controlled run 1.662 kWh, a 6.5% energy reduction, while producing 18.1% more tokens; the two effects compound into the efficiency figures. Energy intensity falls from 1.026 to 0.813 Wh per thousand tokens, a 20.8% improvement, and at the industrial electricity price of \$0.08/kWh the cost of generation falls from \$0.082 to \$0.065 per million tokens. Carbon follows energy at the New York grid intensity of 0.25 kg $CO_2$/kWh, falling from 256.5 to 203.2 grams per million tokens. The absolute magnitudes of a single 500-step run are small by construction; the ratios are the result, because they are scale-free and carry directly to production workloads of the same class.

Figure 8 places the three experiments side by side in the units an operator budgets in. Energy per run reflects duration as much as scale, with the 500-step 7B run at 1.78 kWh and the 200-step 72B run at 1.83 kWh, while mean power per device falls with sharding, from 314.9 W on the monolithic 7B job through 190.7 W at 14B to 147.0 W at 72B, the signature of pipelined execution that Section 6.3 diagnosed. Carbon per run follows energy directly at the regional grid intensity, from 444 grams for the 7B run to 457 grams for the 72B run. Figure 9 presents the four panels of the ledger: violations, tokens, energy intensity, and tokens per megawatt-hour.

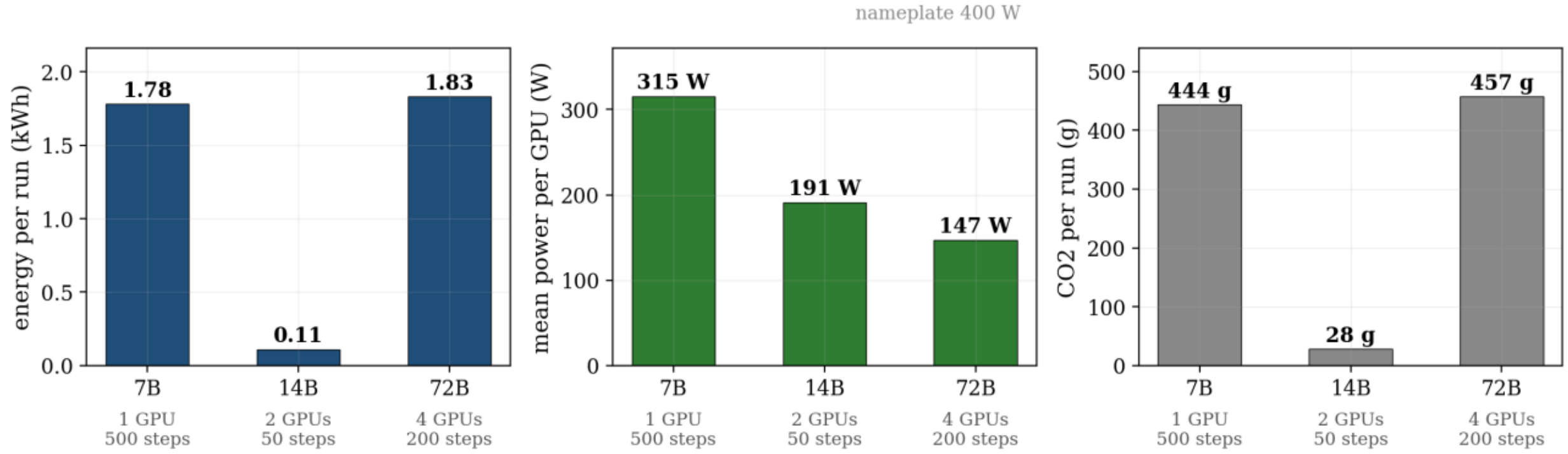

*FIGURE 8. Energy per run, mean power per device, and carbon per run for the three experiments. Per-device power falls with sharding while nameplate remains 400 W per device.*

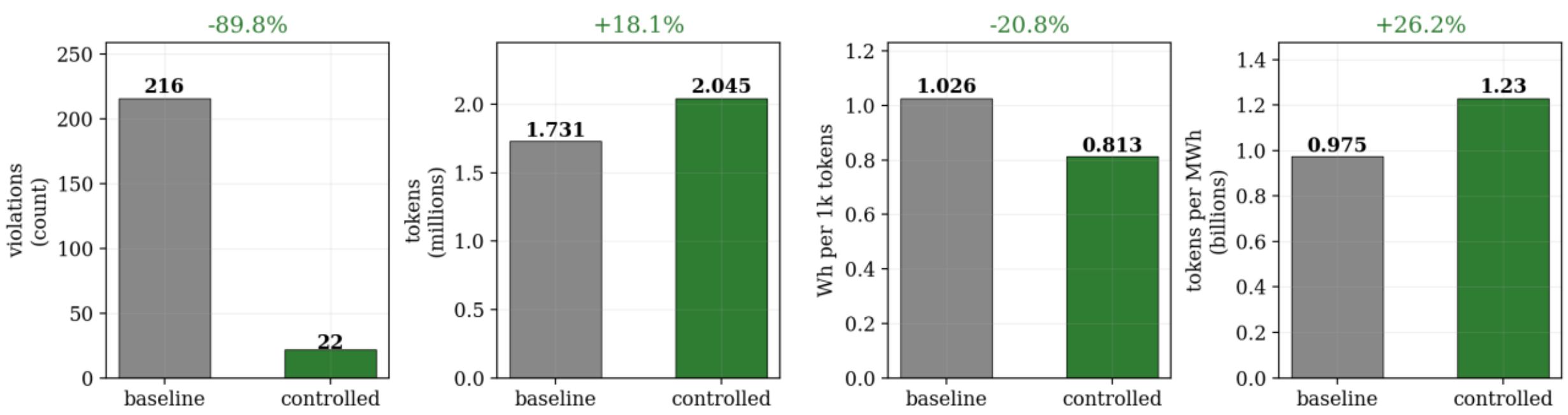

*FIGURE 9. The controller's effect on the 7B evaluation in four panels: cap violations, tokens produced, energy intensity in Wh per thousand tokens, and tokens per megawatt-hour, baseline against controlled.*

***Table 8. Energy, cost, and carbon ledger for the 7B evaluation (500 steps, measured)***

| Quantity | Baseline | Controller | Change |
|---|---|---|---|
| Energy consumed, full timeline (kWh) | 1.777 | 1.662 | −6.5% |
| Tokens produced | 1,731,712 | 2,044,928 | +18.1% |
| Energy intensity (Wh per 1,000 tokens) | 1.026 | 0.813 | −20.8% |
| Generation cost ($ per million tokens) | 0.082 | 0.065 | −20.8% |
| Carbon intensity (g $CO_2$ per million tokens) | 256.5 | 203.2 | −20.8% |
| Tokens per MWh | 0.975 B | 1.230 B | +26.2% |

Projected to facility scale, the energy channel follows one explicit equation,

$$E_{avoided} = C \cdot u \cdot 8760 \cdot f \cdot s \quad (9)$$

with installed capacity C in MW, utilization u = 0.70, controllable-workload fraction f = 0.50, and s = 0.208 the fixed-output energy saving, the framing in which the controlled system produces the same tokens for less energy. For a 100 MW facility this yields approximately 63,800 MWh, $5.1 million, and 15,900 tonnes of $CO_2$ per year, scaling linearly with capacity. These projections deliberately count only the energy-efficiency channel of Table 8; the capacity channel of Section 8, which operates on what a facility must build rather than on what it consumes, is accounted separately in Section 9.

## 7. What a violation is: measurement-window analysis

A question that should precede any control effort: violations at what measurement window? Electrical infrastructure does not respond to half-second samples. Thermal-magnetic breakers integrate over tens of seconds before tripping at moderate overloads, and utility demand charges are computed on five-to-fifteen-minute averages. Recomputing the 72B record as a rolling mean over increasing windows collapses the violation rate from 23.6% at the instantaneous half-second resolution, through 10.8% at ten seconds and 1.6% at thirty seconds, to exactly zero at five minutes, with no controller involved at all. The composed fleet of Section 8 behaves identically, falling from 9.1% instantaneous to zero at every window of thirty seconds and longer. Three distinct statements must therefore be kept apart: the original BF16 72B transients fall to 1.6% at thirty seconds and vanish by five minutes; the composed fleet is at zero for thirty seconds and longer; and the live occupancy controller of Section 6.6 records 2.27% against a budget explicitly constructed on the thirty-second window, a different quantity that no windowing argument erases. Figure 10 plots the first two series.

The analysis divides violations into two classes with opposite properties, and the division organizes everything else in the paper. Plateau violations, sustained draw above the limit as in the 7B condition whose 43.2% rate survives any averaging, are the class that trips protection and incurs demand charges, and they are precisely the class the controller of Section 6.1 eliminates. Transient violations, the sub-second pipeline overlaps of the multi-GPU condition, are mostly averaged away at the integration times infrastructure actually uses, reaching zero by the five-minute window, require no in-loop control, and would fall to millisecond-class firmware actuators in the unlikely event that any facility metric registered them. On this analysis, the three controllers of Section 6.3 were solving a problem that does not exist at infrastructure-relevant timescales, and the practical import of the multi-GPU nulls is smaller than they first appear: in-loop control fails exactly where it is not needed.

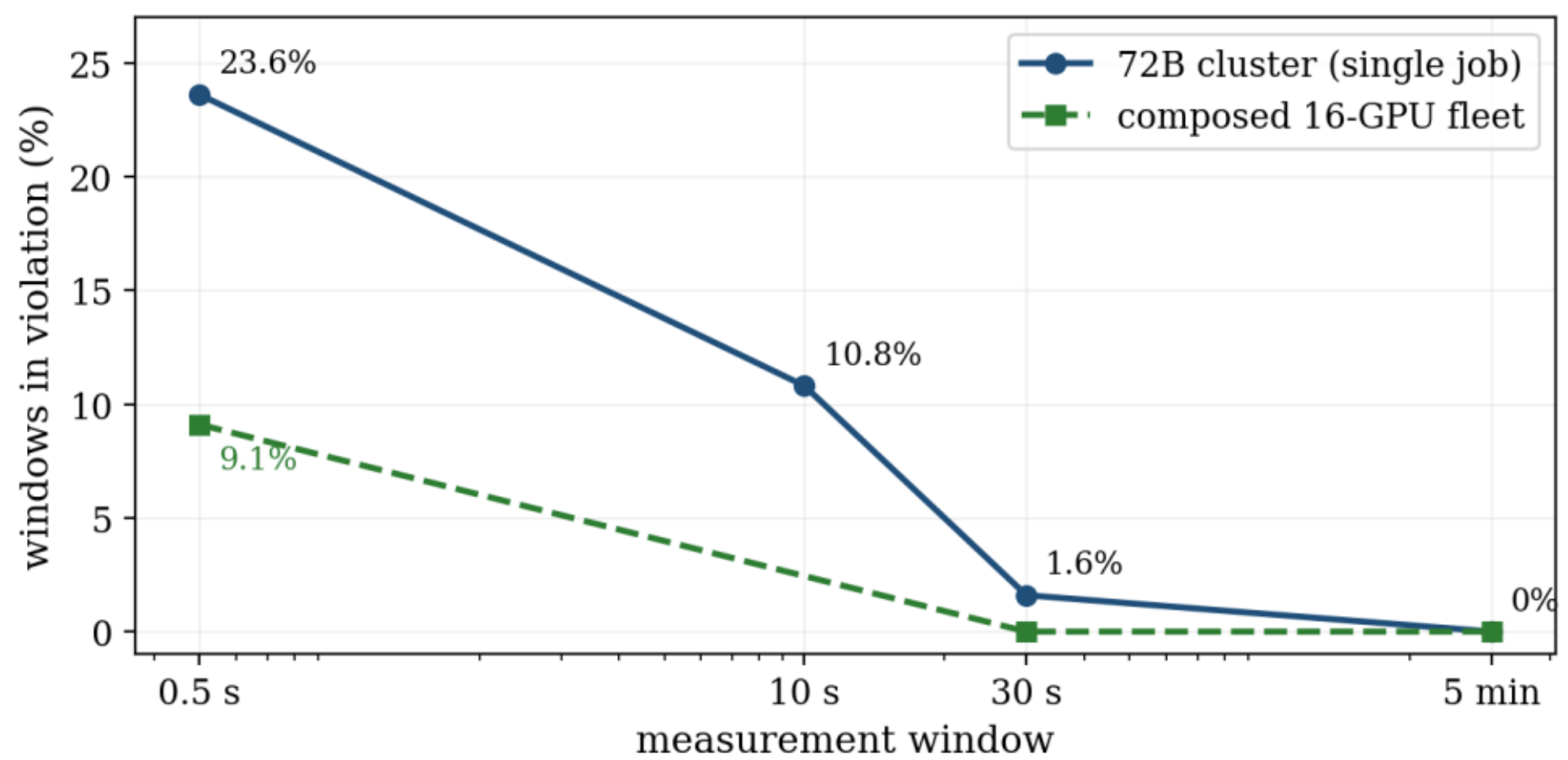

*FIGURE 10. Violation rate against measurement window from half a second to five minutes, for the 72B cluster (both curves labeled) and the composed fleet. The single-job rate falls from 23.6% to 1.6% at thirty seconds and to zero at five minutes; the fleet is at zero for every window of thirty seconds and longer.*

## 8. From traces to facilities: the provisioning gap

The final study composes the measured traces into a facility-scale question: how much electrical capacity does a mixed training fleet actually require? We assemble a representative fleet of two 72B jobs on four accelerators each, two 14B jobs on two each, and four 7B jobs on one each, sixteen A100s in total with an aggregate nameplate of 6.4 kW. Each job's power over time is its real recorded trace tiled over a common horizon; nothing about per-job behavior is modeled, in observance of the simulation-to-hardware finding of Section 6.3. Four operating strategies are compared: naive simultaneous starts representing worst-case phase alignment; random start offsets, which is what job queues produce by accident; phase-staggered starts, in which a greedy scheduler selects each job's offset to minimize the running thirty-second peak; and staggering combined with job-level control, in which the measured controller effect of Section 6.1 is applied to the elastic single-accelerator jobs only. The decision metric is the thirty-second rolling peak, the quantity protection responds to under Section 7, and the derived provisioning fraction, that peak divided by nameplate.

Table 9 reports the outcome. Aggregation does most of the work: the fleet's peak-to-mean ratio is 1.12, against 2.1 for the largest single job, which is statistical multiplexing operating exactly as utility diversity factors predict. Scheduling and job-level control together remove a further six percentage points, bringing the requirement to half the nameplate. Across two independent trace compositions the final figure ranged from 45 to 50% of nameplate, and we report the conservative bound. The claim, stated precisely, is that for this fleet mix, measured behavior supports approximately twofold oversubscription of nameplate capacity with zero violations at any measurement window of thirty seconds or longer. Figure 11 gives the summary view.

***Table 9. Fleet operating strategies (16 GPUs, 6.4 kW nameplate, real traces)***

| Strategy | Mean (kW) | Instantaneous peak (kW) | 30 s peak (kW) | Provisioning fraction |
|---|---|---|---|---|
| Naive (aligned starts) | 2.95 | 4.93 | 3.59 | 56% |

| Strategy | Mean (kW) | Instantaneous peak (kW) | 30 s peak (kW) | Provisioning fraction |
|---|---|---|---|---|
| Random starts | 2.95 | 4.13 | 3.44 | 54% |
| Phase-staggered | 2.95 | 4.10 | 3.38 | 53% |
| Staggered with job control | 2.95 | 4.10 | 3.17 | 50% |

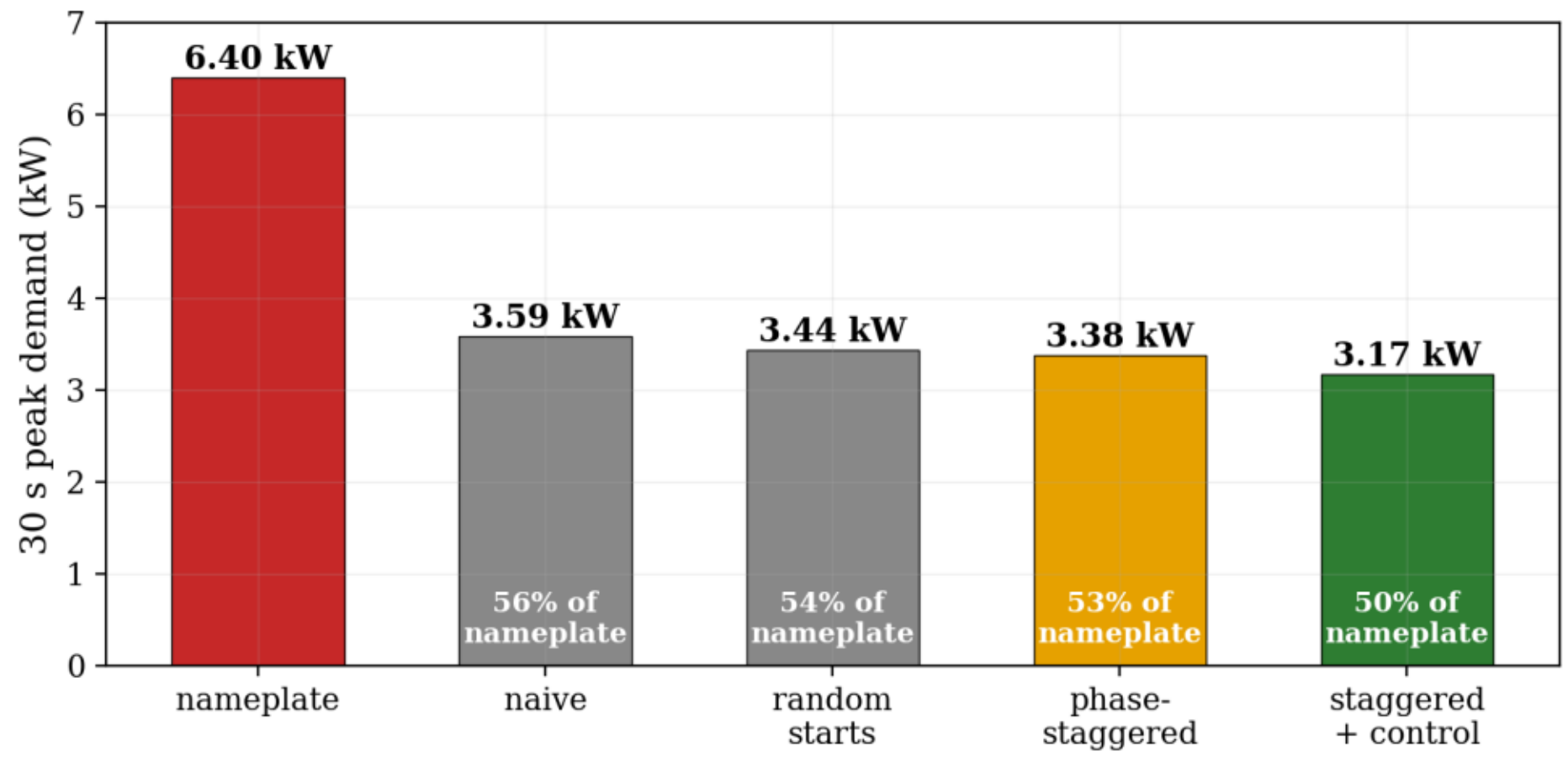

*FIGURE 11. The provisioning gap. Nameplate capacity of the sixteen-GPU fleet against its measured thirty-second peak demand under the four operating strategies.*

## 9. Discussion

### 9.1 Economic and carbon implications

Electrical capacity is the long-lead, high-cost input to an AI facility. Published estimates place construction plus grid interconnection on the order of four million dollars per megawatt in current United States markets [3, 35], and interconnection timelines run to years [5]. At the conservative provisioning fraction of 50%, a facility designed around 100 MW of accelerator nameplate requires roughly 50 MW of provisioned capacity, which corresponds to on the order of two hundred million dollars of avoided capital and several years of queue position per 100 MW, scaling linearly with facility size. A second and separable channel operates on energy rather than capacity: the measured 26.2% gain in tokens per megawatt-hour converts directly into output per unit of electricity on the jobs the controller governs. Applied illustratively to New York State's multi-gigawatt interconnection pipeline of proposed large loads, which includes data centers among other large facilities [36], and treating the AI-attributable share as the scenario quantity, the two channels compound. The capacity channel, at the 50% provisioning fraction, implies on the order of 2,300 MW of avoidable new construction and roughly 9.2 billion dollars of capital. The efficiency channel, computed with the same explicit equation as Section 6.9 (capacity times 70% utilization times 8,760 hours times a 50% controllable fraction times the 20.8% fixed-output saving), corresponds to approximately 2.9 TWh, 235 million dollars, and 733,000 tonnes of $CO_2$ avoided annually across the pipeline at the regional grid intensity, equivalent to removing roughly 159,000 passenger cars from the road. These are scenario quantities whose external inputs, the cost per megawatt, queue rates, pipeline size, electricity price, and grid intensity, are stated and substitutable; the measured quantities they multiply are the provisioning fraction and the efficiency gain.

The results also bear on grid integration beyond any single facility. Utility research programs are assembling the evidence base for datacenter flexibility [37], and the load-growth literature identifies curtailable large loads as the near-term path for absorbing AI demand [10, 11]. What those programs have lacked is a demonstrated mechanism by which a training facility sheds load without destroying the jobs it runs. Section 6.1 supplies that mechanism: a controller that substantially reduces violations while increasing output, in trace-replay evaluation, which makes training a dispatchable load in the demand-response sense. Whether artificial intelligence ultimately accelerates or delays decarbonization has been argued to depend on precisely such deployment choices [38].

### *9.2 A deployment architecture implied by the measurements*

The actuator hierarchy of Section 6.3.1 dictates a three-layer control architecture in which each layer is matched to the disturbance timescale the data revealed, and the assignment of actuator to layer follows from the measured authority and response-time constraints rather than from convenience. The first layer is job-level reinforcement-learning control; for elastic single-device jobs it acts once per training step and is the mechanism validated in Section 6.1, and for sharded jobs it acts at generation-batch boundaries on the occupancy actuator and is the mechanism validated live in Section 6.6. The second layer is the industry-standard firmware power cap acting in milliseconds, held in reserve for transient events, which Section 7 shows infrastructure metrics never see; reinforcement-learned dynamic cap-setting is a natural extension that requires the host-level access of an operator deployment. The third layer is a fleet scheduler acting over minutes, allocating power budgets and staggering job phases, which is the strategy measured in Section 8; the scheduler retains responsibility for jobs that expose no runtime actuator, while sharded jobs with a generation phase are additionally governable at the job level through occupancy (Section 6.6). The stack is software-only on existing hardware.

### *9.3 What reinforcement learning contributes*

The evidence that learning pays is threefold: roughly four times the heuristic's violation reduction at identical sensors and actuator; budget behavior no static rule reproduces, converting constraint slack into throughput where loose and enforcing to target where binding; and zero-shot transfer of one cap-normalized policy across three hardware scales in trace simulation, a transfer whose live-hardware analogue is the budget renormalization of Section 6.6 rather than a demonstrated policy transfer. In deployment terms, learning upgrades the load from cappable to dispatchable: a static cap sacrifices throughput at all times to be safe at some times, while the learned policy spends the power budget where it purchases the most output.

## 10. Limitations and validation path

The claims are bounded by the following limitations, which converge on a single validation step. The occupancy-controller results of Sections 6.4 through 6.6 use an AWQ-quantized generation proxy for the rollout phase rather than the BF16 GRPO training loop, disclosed and justified by the rollout's 99.9% share of measured wall time but not identical to it. The live controller missed the pre-specified 1% violation bar (2.27%), its advantage over the adaptive heuristic varied with operating condition, and one of three training seeds passed validation; the learned result is a best average operating point, not uniform dominance. The headline controller result of Section 6.1 is an evaluation of the trained policy against the full measured 7B trace: real hardware telemetry, but not a live in-loop hardware run, whereas live in-loop operation is verified mechanically at the 72B scale, exactly where the original group-size actuator lacks authority. The 50% provisioning fraction is composed from real traces rather than measured on an operated fleet. The headline runs are single runs with large effect sizes, while the multi-GPU null is the replicated result. The economic conversions rest on published cost and queue figures rather than procurement data. And the experimental power caps are constructed conditions mapped to oversubscription scenarios rather than to a specific facility tariff.

The validation step that addresses these is an instrumented pilot on one operating mixed fleet: this paper's telemetry stack deployed alongside the operator's revenue-grade metering, the fleet scheduler applied to all jobs and job-level controllers to the elastic ones, for thirty to sixty days at an estimated cost of fifty thousand dollars, with a pre-specified success criterion of at least a 30% reduction in required provisioning relative to nameplate at zero violations on any thirty-second window of the operator's own metering. The deliverable, an operator-verified provisioning fraction, would convert the central estimate of this paper into a planning figure and simultaneously upgrade the controller evidence from trace evaluation to live fleet operation.

## 11. Conclusion

Reinforcement-learning post-training, the dominant workload of the current era of model development, has measurable and exploitable power structure. A learned controller over GRPO's generation parameters, trained and evaluated on real A100 telemetry, holds training inside a power budget while increasing output: 89.8% fewer violations, 18.1% more tokens, 26.2% more tokens per megawatt-hour. The same campaign locates and then crosses the boundary of the approach. In-loop group-size control loses authority on sharded multi-GPU jobs; the actuator sweep shows generation concurrency retains 17 to 22% authority there; and a controller rebuilt on the occupancy actuator improved the live 72B rollout-generation throughput and constraint trade-off across three hardware replications, 35.7% more tokens than the static safe configuration and 87% fewer violations than uncontrolled full occupancy. Residual sub-second transients are largely averaged out at infrastructure timescales and gone entirely at five minutes. Composed to fleet scale, the traces show mixed training fleets requiring roughly half their nameplate capacity. For the measured fleet mix, these results indicate that provisioning to nameplate carries roughly twice the capacity that controlled, measured demand requires, a conclusion whose generalization beyond similar workload mixes awaits operator validation; a gap worth on the order of two hundred million dollars and years of interconnection queue per hundred megawatts, and that a fifty-thousand-dollar pilot separates this measured estimate from an operator-verified standard. Measured and controlled, AI training is not the inflexible load the grid fears. It is among the most schedulable large loads the grid has been offered.

## References

[1] A. Shehabi, S. J. Smith, A. Hubbard, A. Newkirk, N. Lei, M. A. B.
Siddik, B. Holecek, J. Koomey, E. Masanet, and D. Sartor, "2024 United States Data Center Energy Usage Report," Lawrence Berkeley National Laboratory, LBNL-2001637, 2024.

[2] International Energy Agency, "Energy and AI," IEA Special Report,

Paris, 2025.

[3] Goldman Sachs Global Investment Research, "Generational Growth:
AI, Data Centers and the Coming US Power Demand Surge," 2024.

[4] International Energy Agency, "Electricity 2025: Analysis and
Forecast to 2027," IEA, Paris, 2025.

[5] J. Rand, N. Manderlink, W. Gorman, R. Wiser, J. Seel, J. M. Kemp,
S. Jeong, and F. Kahrl, "Queued Up: 2024 Edition, Characteristics of Power Plants Seeking Transmission Interconnection," Lawrence Berkeley National Laboratory, 2024.

[6] Federal Energy Regulatory Commission, "Improvements to Generator
Interconnection Procedures and Agreements," Order No. 2023, Docket RM22-14, 2023.

[7] Electric Power Research Institute, "Powering Intelligence:
Analyzing Artificial Intelligence and Data Center Energy Consumption," EPRI White Paper 3002028905, 2024.

[8] A. de Vries, "The growing energy footprint of artificial
intelligence," Joule, vol. 7, no. 10, pp. 2191-2194, 2023.

[9] Uptime Institute, "Global Data Center Survey 2024," Uptime
Institute Intelligence, 2024.

[10] T. Norris, T. Profeta, D. Patiño-Echeverri, and A. Cowie-Haskell,
"Rethinking Load Growth: Assessing the Potential for Integration of Large Flexible Loads in US Power Systems," Nicholas Institute for Energy, Environment and Sustainability, Duke University, 2025.

[11] U.S. Department of Energy, "Recommendations on Powering
Artificial Intelligence and Data Center Infrastructure," Secretary of Energy Advisory Board, 2024.

[12] P. Patel, E. Choukse, C. Zhang, Í. Goiri, B. Warrier, N.
Mahalingam, and R. Bianchini, "Characterizing Power Management Opportunities for LLMs in the Cloud," in Proc. 29th ACM International Conference on Architectural Support for Programming Languages and Operating Systems (ASPLOS), 2024.

[13] P. Patel, E. Choukse, C. Zhang, A. Shah, Í. Goiri, S. Maleki, and
R. Bianchini, "Splitwise: Efficient Generative LLM Inference Using Phase Splitting," in Proc. 51st International Symposium on Computer Architecture (ISCA), 2024.

[14] J. Stojkovic, C. Zhang, Í. Goiri, J. Torrellas, and E. Choukse,
"DynamoLLM: Designing LLM Inference Clusters for Performance and Energy Efficiency," in Proc. IEEE International Symposium on High-Performance Computer Architecture (HPCA), 2025.

[15] S. Samsi, D. Zhao, J. McDonald, B. Li, A. Michaleas, M. Jones, W.

Bergeron, J. Kepner, D. Tiwari, and V. Gadepally, "From Words to Watts: Benchmarking the Energy Costs of Large Language Model Inference," in Proc. IEEE High Performance Extreme Computing Conference (HPEC), 2023.

[16] J. You, J.-W. Chung, and M. Chowdhury, "Zeus: Understanding and Optimizing GPU Energy Consumption of DNN Training," in Proc. 20th USENIX Symposium on Networked Systems Design and Implementation (NSDI), 2023.

[17] J.-W. Chung, Y. Gu, I. Jang, L. Meng, N. Bansal, and M. Chowdhury, "Perseus: Removing Energy Bloat from Large Model Training," in Proc. 30th ACM Symposium on Operating Systems Principles (SOSP), 2024.

[18] A. S. Luccioni, S. Viguier, and A.-L. Ligozat, "Estimating the Carbon Footprint of BLOOM, a 176B Parameter Language Model," Journal of Machine Learning Research, vol. 24, no. 253, pp. 1-15, 2023.

[19] A. S. Luccioni, Y. Jernite, and E. Strubell, "Power Hungry Processing: Watts Driving the Cost of AI Deployment?" in Proc. ACM Conference on Fairness, Accountability, and Transparency (FAccT), 2024.

[20] E. Choukse et al., "Power Stabilization for AI Training Datacenters," arXiv:2508.14318, 2025.

[21] R. Zheng, S. Dou, S. Gao, Y. Hua, W. Shen, B. Wang, et al., "Secrets of RLHF in Large Language Models Part I: PPO," arXiv:2307.04964, 2023.

[22] M. Towers, A. Kwiatkowski, J. Terry, J. U. Balis, G. de Cola, T. Deleu, et al., "Gymnasium: A Standard Interface for Reinforcement Learning Environments," arXiv:2407.17032, 2024.

[23] A. Raffin, A. Hill, A. Gleave, A. Kanervisto, M. Ernestus, and N. Dormann, "Stable-Baselines3: Reliable Reinforcement Learning Implementations," version 2.x, 2023. Software.

[24] S. Gu, L. Yang, Y. Du, G. Chen, F. Walter, J. Wang, and A. Knoll, "A Review of Safe Reinforcement Learning: Methods, Theories and Applications," IEEE Transactions on Pattern Analysis and Machine Intelligence, 2024.

[25] J. Ji, B. Zhang, J. Zhou, X. Pan, W. Huang, R. Sun, et al., "Safety Gymnasium: A Unified Safe Reinforcement Learning Benchmark," in NeurIPS Datasets and Benchmarks Track, 2023.

[26] I. Kulikov, C. Whitehouse, T. Wu, Y. Nie, S. Saha, E. Helenowski, et al., "Autodata: An Agentic Data Scientist to Create High Quality Synthetic Data," arXiv:2606.25996, Meta FAIR, 2026.

[27] OpenAI, "GPT-4 Technical Report," arXiv:2303.08774, 2023.

[28] R. Rafailov, A. Sharma, E. Mitchell, S. Ermon, C. D. Manning, and

C. Finn, "Direct Preference Optimization: Your Language Model is Secretly a Reward Model," in Advances in Neural Information Processing Systems 36 (NeurIPS), 2023.

[29] Z. Shao, P. Wang, Q. Zhu, R. Xu, J. Song, X. Bi, et al., "DeepSeekMath: Pushing the Limits of Mathematical Reasoning in Open Language Models," arXiv:2402.03300, 2024.

[30] DeepSeek-AI, "DeepSeek-R1: Incentivizing Reasoning Capability in LLMs via Reinforcement Learning," arXiv:2501.12948, 2025.

[31] NVIDIA Corporation, "NVIDIA A100 Tensor Core GPU Architecture," Technical Documentation, v2.1, 2023.

[32] L. von Werra, Y. Belkada, L. Tunstall, E. Beeching, T. Thrush, N. Lambert, et al., "TRL: Transformer Reinforcement Learning," version 0.9+, Hugging Face, 2024. Software.

[33] NVIDIA Corporation, "NVIDIA Management Library (NVML) API Reference," Developer Documentation, 2024.

[34] Qwen Team, "Qwen2.5 Technical Report," arXiv:2412.15115, 2025.

[35] McKinsey & Company, "AI power: Expanding data center capacity to meet growing demand," 2024.

[36] New York Independent System Operator, "2024 Load & Capacity Data Report (Gold Book)," Rensselaer, NY, 2024.

[37] Electric Power Research Institute, "DCFlex: Data Center Flexibility Initiative," EPRI, 2024.

[38] A. Luers, J. Koomey, E. Masanet, N. Gillett, T. Hertel, J. Jenkins, and D. Rejeski, "Will AI accelerate or delay the race to net-zero emissions?" Nature, vol. 628, pp. 718-720, 2024.